\documentclass[journal]{IEEEtran}

\usepackage{cite}
\usepackage{graphicx}
\usepackage[lowtilde]{url}
\usepackage{pgfplots}
\pgfplotsset{compat=newest}
\usetikzlibrary{plotmarks}
\usepackage{grffile}
\usepackage{amsmath}
\usepackage{multirow}
\usepackage{enumitem}
\usepackage[euler]{textgreek}
\usepackage{booktabs} 
\usepackage{array}
\usepackage{verbatim}
\usepackage{amsfonts,amssymb}
\usepackage{algorithm}
\usepackage{algorithmicx}
\usepackage{algpseudocode}
\usepackage{colortbl}
\usepackage{color}
\definecolor{mygray}{gray}{.9}
\usepackage{diagbox} 
\usepackage{graphicx}
\usepackage{subfigure} 
\usepackage{bm}
\usepackage{caption}   

\makeatletter
\long\def\@makecaption#1#2{\ifx\@captype\@IEEEtablestring%
\footnotesize\begin{center}{\normalfont\footnotesize #1}\\
{\normalfont\footnotesize\scshape #2}\end{center}%
\@IEEEtablecaptionsepspace
\else
\@IEEEfigurecaptionsepspace
\setbox\@tempboxa\hbox{\normalfont\footnotesize {#1.}~~ #2}%
\ifdim \wd\@tempboxa >\hsize%
\setbox\@tempboxa\hbox{\normalfont\footnotesize {#1.}~~ }%
\parbox[t]{\hsize}{\normalfont\footnotesize \noindent\unhbox\@tempboxa#2}%
\else
\hbox to\hsize{\normalfont\footnotesize\hfil\box\@tempboxa\hfil}\fi\fi}
\makeatother

\newcommand{\RN}[1]{%
  \textup{\uppercase\expandafter{\romannumeral#1}}%
}

\newcolumntype{C}[1]{>{\centering\let\newline\\\arraybackslash\hspace{0pt}}m{#1}}

\begin{document}
\title{Progressive Knowledge Transfer Based on Human Visual Perception Mechanism for Perceptual Quality Assessment of Point Clouds}
\author{Qi~Liu,~Yiyun~Liu,~Honglei~Su,~\IEEEmembership{Member,~IEEE,}~Hui~Yuan,~\IEEEmembership{Senior Member,~IEEE,} and~Raouf~Hamzaoui,~\IEEEmembership{Senior Member,~IEEE}
\thanks{ 

Qi Liu and Honglei Su are with the College of Electronic Information, Qingdao University, Qingdao, 266071, China (email: sdqi.liu@gmail.com, suhonglei@qdu.edu.cn, chentx3854@gmail.com).

Yiyun Liu is with the School of Electrical and Computer Engineering, New York University, US. (email: yiyun.liu@nyu.edu).

Hui Yuan is with the School of Control Science and Engineering, Shandong University, Ji'nan, 250061, China (e-mail: huiyuan@sdu.edu.cn).

Raouf Hamzaoui is with the School of Engineering and Sustainable Development, De Montfort University, Leicester LE1 9BH, U.K. (e-mail: rhamzaoui@dmu.ac.uk).

Corresponding author: Honglei Su.
}
}

\markboth{IEEE Transactions on Multimedia}%
{Shell \MakeLowercase{\textit{et al.}}: Bare Demo of IEEEtran.cls for Journals}

\maketitle

\begin{abstract} With the wide applications of colored point cloud in many fields, point cloud perceptual quality assessment plays a vital role in the visual communication systems owing to the existence of quality degradations introduced in various stages. However, the existing point cloud quality assessments ignore the mechanism of human visual system (HVS) which has an important impact on the accuracy of the perceptual quality assessment. In this paper, a progressive knowledge transfer based on human visual perception mechanism for perceptual quality assessment of point clouds (PKT-PCQA) is proposed. The PKT-PCQA merges local features from neighboring regions and global features extracted from graph spectrum. Taking into account the HVS properties, the spatial and channel attention mechanism is also considered in PKT-PCQA. Besides, inspired by the hierarchical perception system of human brains, PKT-PCQA adopts a progressive knowledge transfer to convert the coarse-grained quality classification knowledge to the fine-grained quality prediction task. Experiments on three large and independent point cloud assessment datasets show that the proposed no reference PKT-PCQA network achieves better of equivalent performance comparing with the state-of-the-art full reference quality assessment methods, outperforming the existed no reference quality assessment network. 
\end{abstract}


\IEEEpeerreviewmaketitle

\section{Introduction}\label{sec:introduction}

\IEEEPARstart {T}{hanks} to the rapid development of three-dimensional (3D) data acquisition technologies, colored point clouds are now readily available and popular. As an effective representation of 3D visual contents, a point cloud consists of a set of points, each of which contains geometric coordinates and other attributes such as color~\cite{liu2020model}. Point clouds have been widely used in automatic driving~\cite{9257015}~\cite{gu20193d}, immersive telepresence~\cite{9555219} and so on. However, since point clouds record the omnidirectional detail of objects and scenes, they usually need large storage space and very high transmission bandwidth in practical applications. Hence, various 3D processing algorithms were proposed to adapt to specific needs using processing operations such as simplification and compression, which inevitably cause damage to the visual quality of 3D models~\cite{zhang2022no}~\cite{liu2019comprehensive}. The Moving Picture Experts Group (MPEG) has been developing two point cloud compression standards: geometry-based point cloud compression (G-PCC)~\cite{gpcc} and video-based point cloud compression (V-PCC)~\cite{vpcc}. Therefore, developing point cloud quality assessment (PCQA) techniques can help to understand the distortion and carry out the quality optimization for point cloud. 

Since the human visual system (HVS) is the ultimate receiver of 3D point clouds in most applications, the only reliable method to assess the point cloud quality perceived by a human observer is to ask human subjects for their opinion~\cite{wu2021subjective}. Subjective PCQA is impractical for most applications due to the human involvement in the process. Alexiou~\textit{et al.}~\cite{alexiou2019comprehensive} focused on the evaluation of test conditions defined by MPEG for core experiments including 99 distorted point clouds. Javaheri~\textit{et al.}~\cite{javaheri2020point} encoded six original voxelized point clouds from MPEG by three codecs with different quality generating 54 distorted point clouds. Yang~\textit{et al.}~\cite{yang2020predicting} produced a point cloud subjective evaluation dataset, denoted as SJTU-PCQA, with 420 samples at different distortion levels. Additionally, Su~\textit{et al.}~\cite{su2019perceptual} carried out a subjective quality assessment experiment covering 740 distorted point clouds. However, subjective PCQA studies provide valuable data to assess the performance of objective or automatic methods of quality assessment~\cite{su2019perceptual}. Subjective studies also enable improvements in the performance of objective PCQA algorithms toward attaining the ultimate goal of matching human perception. Like image/video quality assessment methods, objective PCQA methods can be classified into three categories: full reference (FR), reduced reference (RR), and no reference (NR) methods~\cite{alexiou2021perceptual}. To evaluate the quality of a distorted point cloud, FR methods use the pristine uncompressed point cloud as a reference, while RR methods only require statistical features that are extracted from the reference point cloud. On the other hand, NR methods evaluate the quality of the distorted point cloud in the absence of the reference. Although substantial efforts have been made to develop objective PCQA models, the proposed solutions often fail to draw a connection to the HVS or struggle in handling the irregular representation of point clouds~\cite{zhang2022no}. 

In this paper, we propose a progressive knowledge transfer based on a human visual perception mechanism for perceptual quality assessment of point clouds (PKT-PCQA). PKT-PCQA designs an intelligent network model to simulate the process of human visual processing of point cloud information, including the global and local quality mergence, attention mechanism and functional hierarchical architecture of human vision. The main contributions of this paper are as follows:
\begin{enumerate}
\item We develop a perceptual quality assessment model of point clouds. Inspired by the hierarchical architecture for human vision, we exploit the coarse-to-fine learning strategy to transfer the information from a coarse quality classification to fine quality prediction task. Extensive experimental results show that our model is more accurate than existing FR and RR point cloud quality models.
\item We introduce a hierarchical attention mechanism in the 3D domain. The proposed mechanism is a dynamic weight adjustment process based on features of the input point cloud. Since the attention mechanism is very similar to the human selective visual attention mechanism, our model can make a more accurate prediction of the human subjective opinion score than previous models.
\item Inspired by multiple scales of receptive fields of human vision, we propose a novel feature extraction framework to consider the impact of the global and local features on the point cloud quality. The global geometry feature is for key clusters extraction, reducing the complexity of the network computation and increasing the similarity between the network quality score and human perceived quality.
\end{enumerate}

The remainder of this paper is organized as follows. In Section \ref{sec:related work}, we discuss related work. In Section~\ref{sec:proposed}, we propose a quality assessment network that considers the visual perception for point clouds. We conduct experimental results in Section~\ref{sec:Experimental Results and Discussion} and draw conclusions in Section~\ref{sec:conclusion}.

\begin{figure*}[t!]
\centering{\includegraphics[width=1\textwidth]{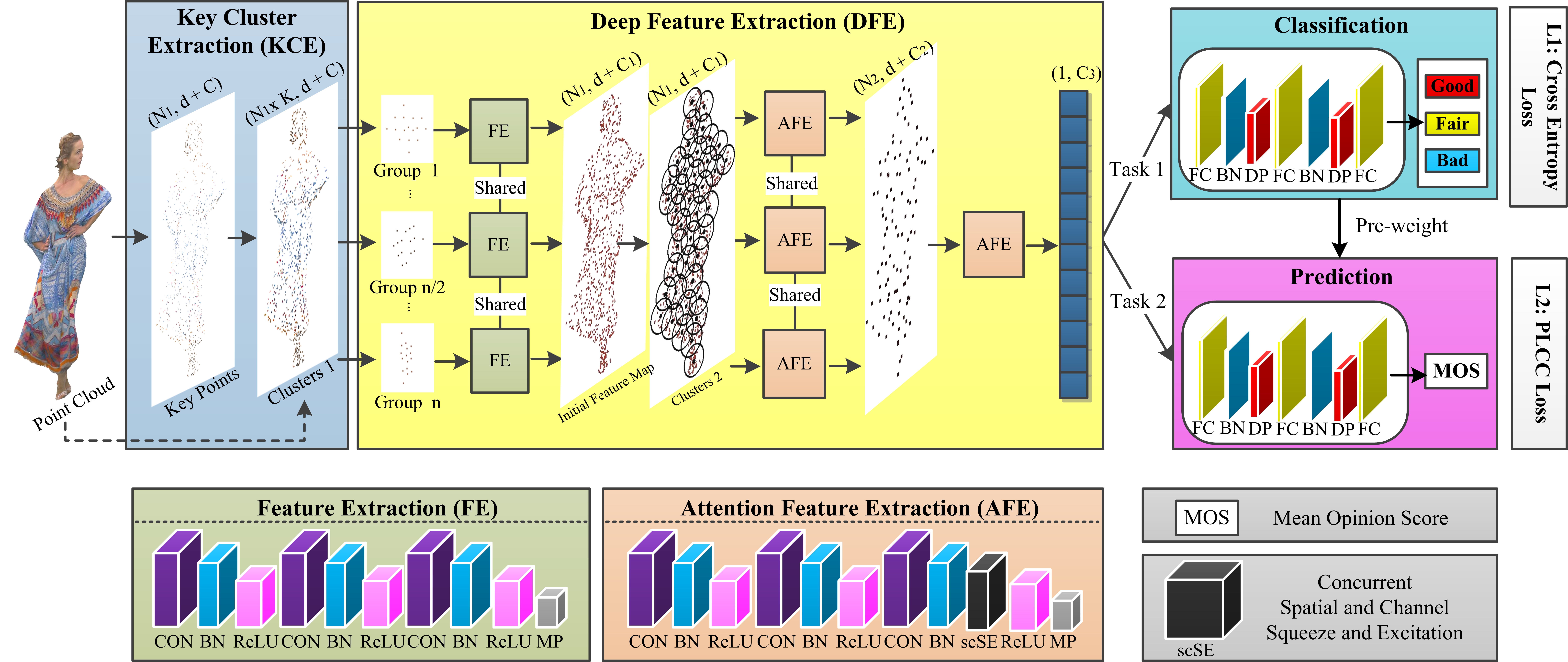}}
\caption{\textbf{PKT-PCQA Architecture.} The classification task and prediction task share the same feature extraction operation, which includes key clusters extraction (KCE) and deep feature extraction (DFE). The point cloud represented in each stage named clusters1 and clusters2 respectively.  $(N_1 \times K, d + C)$ represents $N_1 \times K$ points with $d$-dimension coordinates and $C$-dimension features, where "$N1$" is the number of key points generated from the previous key points extraction operation, and "$K$" is the number of points aggregated in each cluster. In the Feature Extraction (FE) and Attention Feature Extraction (AFE) units, CON denotes convolution layer, BN is the batch normal unit, ReLU is used as activation function, and MP denotes maxpooling. FC is a fully connected module, DP is the dropout unit, and scSE is a concurrent spatial and channel 'squeeze and excitation' attention module.\label{fig:framework}}
\end{figure*}
\section{Related Work}\label{sec:related work}
FR objective quality assessment techniques for colored point clouds can be classified into two categories: point-based metrics and projection-based metrics. Point-based metric identifies the mean squared error between the color information of the points in the original point cloud and the distorted point cloud~\cite{mekuria2017design}. The most representative point-based FR point cloud quality metric is point-to-point PSNR only for geometry or color~\cite{tian2017geometric}~\cite{mekuria2017performance2} , namely $PSNR_{MSE,p2po}$, $PSNR_{HF,p2po}$ and $PSNR_Y$, respectively. In addition, statistics of a variant of the Local Binary Pattern~\cite{9190956}~\cite{diniz2022point}, Perceptual Color Distance Pattern~\cite{diniz2021novel} and Local Luminance Pattern~\cite{9287154} descriptors were introduced to assess the quality of the point clouds. Based on the above pioneer studies, 
Yang~\textit{et al.}~\cite{yang2020inferring} constructed local graphs for both reference and distorted point clouds to compute three moments of color gradients, which are used to obtain similarity index by pooling the local graph significance across all channels, namely $GraphSIM$. Excepted that, the features of every point are commonly used to assess the point cloud quality in the point-based categories. Viola, Subramanyam, and Cesar~\cite{viola2020color} only exploited color histograms and correlograms to estimate the impairment of a distorted point cloud with respect to its reference. However, in $PointSSIM$~\cite{alexiou2020towards} and   $PCQM$~\cite{meynet2020pcqm}, both geometry-based and color-based features were extracted to predict the point cloud quality. Considering the visual masking effect of the geometric information of point cloud and the color perception of human eyes, Hua~\textit{et al.}~\cite{hua2022cpc} used geometric segmentation and color transformation respectively to construct geometric and color features and then to estimate the point cloud quality. Inspired by the point cloud generation process, Xu~\textit{et al.}~\cite{9552578}introduced elastic forces to record the shaping of the point set and used the elastic potential energy difference to quantify the point cloud distortion. Inspired by the research of elastic potential energy, Yang ~\textit{et al.}~\cite{yang2022mped} split the source and target point clouds into multiple neighborhoods and measure the discrepancy between the two points clouds as a multiscale point potential energy discrepancy. The potential energy is defined around the centers of the neighborhoods in a way that reflects the geometry, colour, and contextual information in the point cloud. Lu~\textit{et al.}~\cite{lu2022point} designed a dual-scale 3D-DOG filters to explore the 3D edge information inherent in the PCs, and the corresponding 3D edge similarity of the reference and distorted point clouds is calculated to measure the quality loss of the distorted PCs. Additionally, the projection-based metrics project the 3D point cloud onto a 2D surface generating multiple 2D images of the point cloud. Then the existing image quality assessment methods SSIM~\cite{wang2004image}, MS-SSIM~\cite{wang2003multiscale}, $IW-SSIM_{p}$~\cite{liu2022perceptual} and VIFP~\cite{Sheikh2006image} are used to predict the quality of the point cloud from the average quality of the projection images and the other one is a layered projection-based $LP-PCQM$~\cite{9448078} metric and use local and global statistics of the points to assess the perceived quality from both geometry and color. He~\textit{et al.}~\cite{he2022tgp} obtained texture and geometry projection maps from different perspectives for evaluating the colored point cloud. Freitas~\textit{et al.}~\cite{freitas2022point} mapped attributes from the point clouds onto the folded 2D grid, generating a pure-texture 2D image that contains point cloud texture information. Then, they extracted statistical features from these texture maps using a multi-scale rotation-invariant texture descriptor. Next, they computed the geometrical similarities using geometry-only distances. Finally, they fused the texture and geometrical similarities using a stacked regressor to model the point cloud visual quality.

RR objective quality assessment models commonly extract a small amount of information from the reference and distorted point clouds, and then compare and analyze the extracted feature data to predict the point cloud quality. Viola and Cesar proposed a RR quality metric $PCM\_{RR}$~\cite{viola2020reduced}, and they extracted 21 geometry, normal, and luminance features from the reference and distorted point clouds and built an RR quality metric as a weighted sum of their absolute differences. In the previous work~\cite{liu2021reduced}, we proposed an RR linear model that accurately predicts the perceptual quality of compressed point clouds from the geometry and color quantization step sizes. Besides, we also used the information extracted from a bitstream for real-time and nonintrusive quality monitoring~\cite{liu2022no}, and then the geometry quantization parameter, texture quantization parameter, texture bitrate per pixel are used to estimate the point cloud distortion. 

However, often the reference point cloud is not available. Therefore, developing NR objective quality assessments can further facilitate the application of point cloud in practice. Hua~\textit{et al.}~\cite{9547070} proposed a metric dedicated to characterize the distortion of the distorted point cloud from geometric, color and joint perspectives. Since deep learning has achieved great success in various fields, deep learning-based metrics have also been proposed. Chetouani~\textit{et al.}~\cite{chetouani2021deep} used a two-step procedure to divide the complex point cloud quality assessment task into feature extraction and quality prediction. In the previous work, we proposed a deep learning network using multiview 3D point cloud projection images to predict the quality of the point cloud~\cite{liu2021pqa}. The network applies feature extraction, distortion type identification, and quality vector prediction modules. Liu~\textit{et al.}~\cite{liu2022point} proposed a end-to-end sparse convolutional neural network to estimate the subjective quality of point cloud, which exploited a stack of sparse convolutional layers and residual blocks to extract hierarchical features. Tu~\textit{et al.}~\cite{tu2022v} designed a dual-stream convolutional network from the perspective of global and local feature description to extract texture and geometry features of the distorted point cloud. Regrettably, the most of the existed NR learning-based quality prediction networks ignore the Visual attention mechanism of HVS. Tao~\textit{et al.}~\cite{tao2021point} took into account the attention mechanism of HVS in the spatial pooling module and realized the weighted summation of local regional quality to obtain the final global quality score of colored point clouds. However, Tao he weighted summation of the visual quality scores of all local patches from the color and geometric projection maps. On the one hand, the visual perception of the projected 2D image is still different from the 3D visual perception. On the other hand, the 2D projection image in~\cite{tao2021point} is composed of fragmented projections in multiple areas, which is also very different from the overall perceptual image received by the human eye.

Based on the above analysis, it still requires substantial efforts to develop objective metrics to accurately predict the subjective 3D point cloud quality. On the one hand, it comprehensively extracts point cloud features based on the 3D visual characteristics of HVS; on the other hand, inspired by the processing mechanism of visual information, a reasonable network structure should be designed to exploit the extracted features to predict the point cloud quality.
\section{Proposed Method}\label{sec:proposed}
The overall network architecture is presented in Section III-A. Then, key clusters extraction (KCE), deep feature extraction (DFE), and the progressive prediction mechanism are introduced in the successive sections, respectively.
\subsection{Overview of the Proposed Method}
In our proposed network, we use PointNet\cite{qi2017pointnet++} as the backbone, which is well known for its elegance, simplicity, and excellent generalization. For effective use of key information in point clouds, we propose to extract the key clusters from the input point cloud first. Moreover, to simulate the human perception mechanism of point cloud quality, we add an attention mechanism to the network and introduce a coarse-to-fine progressive perception mechanism. In the key clusters extraction stage, we consider global geometry characteristics of the point clouds. During the deep feature extraction stage, we draw on the global and local information of the point cloud and add an attention mechanism to the last two feature extraction steps. The coarse-to-fine perception mechanism runs through the whole point cloud quality prediction network, and the overall architecture of our network is illustrated in Fig.~\ref{fig:framework}

\subsection{Key Clusters Extraction}
Due to its large size, it is difficult to use the point cloud object directly as the input to the no-reference network. We use key points and local clusters to represent the key clusters from the original point cloud and then feed these key clusters into the no-reference network, which solves the network process limitation without sacrificing the effective information of the original point cloud.

(1) Key Points Extraction

According to research in the field of neuroscience, the structure characteristics of objects are very important for human eye perception. For 2D image quality assessment, structural features have been widely used~\cite{xue2013gradient}~\cite{ni2017esim}~\cite{fu2018screen}~\cite{ni2018gabor}~\cite{yang2019modeling}. The full-reference point cloud quality evaluation model employing object structural features~\cite{yang2020inferring} has also shown promising results. For a 3D object, the general architecture can be described by key point roughly, which can create edges, contours, and skeletons, are usually found in the high spatial-frequency range. In this paper, we use a high-pass graph filtering method ~\cite{chen2017fast} to extract the key points from the input point cloud. 

Suppose a point cloud has $N$ points with three-dimensional coordinates and
three-dimensional color attributes, we first construct a general graph of a point cloud by encoding the local geometry information. Suppose a point cloud has $N$ points with three-dimensional coordinates and three-dimensional color attributes. Then, each point corresponds to the vertex of graph, and each effective connection between two points having positive weight is referred to as edge. The edge weight is defined through an adjacency matrix $\mathbf{W} \in \mathbb{R}^{N \times N}$ as (\ref{W}), where $\mathbf{W}_{\bm{\bm{\vec{x}_i}},\bm{\vec x_j}} $ represents the edge weight between two points $\bm{\bm{\vec{x}_i}}$ and $\bm{\vec x_j}$ using the geometric distance as 
\begin{align}
\mathbf{W}_{\bm{\bm{\vec{x}_i}},\bm{\vec x_j}} 
= \begin{cases} 
e^{-\frac{\parallel \bm{\vec{x}_{i}^{o}}-\bm{\vec{x}_{j}^{o}}\parallel_2^2}{\sigma^2}} \quad \text{if}\parallel \bm{\vec{x}_{i}^{o}}-\bm{\vec{x}_{j}^{o}}\parallel < \tau \\
0\qquad\qquad\qquad\text{otherwise}
\end{cases}
\label{W}
\end{align}
Here $\bm{\vec{x}_{i}^{o}}, \bm{\vec{x}_{j}^{o}}$ are 3D coordinates of points $\bm{\bm{\vec{x}_i}},\bm{\vec x_j}$, $\sigma$ is the variance of graph nodes, and  $\tau$ is a Euclidean distance threshold in clustering neighbor points into the same graph. Given this graph, the location and color information of point clouds are called $graph$  $signals$. Then, we take the graph signal as input and produce another graph signal as an output by graph filter. Let $\bm{\vec{x}_{org}}$ represent the original point cloud, we get a geometric key point $\bm{\vec{x}_{key}}$, after applying a high-pass filter $\Psi(\cdot)$,

\begin{align}
\bm{\vec{x}_{key}}=\left \lfloor \Psi(\bm{\vec{x}_{org}},L) \right \rfloor_{\beta}\in \mathbb{R}^{\beta \times 6}, \beta \ll N,
\label{keypoints}
\end{align}
where $L$ is the length of filter $\Psi(\cdot)$, and $\left \lfloor \cdot \right \rfloor_\beta$ is to intercept the first $\beta$ points as the number of key point, which is sorted from large to small by the calculation result of $\Psi(\cdot)$. Typically, the number of key points is less than the total number, this is to say that $\beta$ is less than $N$. The operation of filter $\Psi(\cdot)$ can be expressed as 
\begin{align}
\Psi(\bm{\vec{x}_i}) = \parallel h(\mathbf{A})\bm{\vec{x}_i^{o}} \parallel_2^2,
\label{frequency}
\end{align}
\begin{align}
h(\mathbf{A}) = \begin{matrix}\sum_{l=0}^{L-1} h_l \mathbf{A}^l\end{matrix} 
\label{filter}
\end{align}
where $h_l$ is $l$-th coefficient, and $L$ is the length of graph filter. $\mathbf{A} \in \mathbb{R}^{N \times N}$ is the most elementary nontrivial graph filter, and here we set $\mathbf{A} = \mathbf{D}^{-1}\mathbf{W}$, where $\mathbf{D} =diag(d_1,...,d_N)\in \mathbb{R}^{N \times N}$ with $d_i = \begin{matrix}\sum_{\bm{\vec{x}_i}} \mathbf{W}_{\bm{\vec{x}_i},\bm{\vec{x}_j}}\end{matrix}$. From a graph filtering perspective, we can also refer to $\bm{\vec{x}_{i}^{o}}$ as the input graph signal, and $\Psi(\bm{\vec{x}_i})$ denote the probability value of this point becoming an edge. This probability is then used as the measurement to decide whether associated point can be selected as the key points.

(2) Key Clusters Formation

The capacity of the model to detect subtle features can be improved by capturing local information. This is achieved by a simple clustering algorithm, using the k-nearest neighbors (k-NN) algorithm within a variable radius centered on key points. Here we refer to it collectively as the  center-neighbor-segmentation (CNS) algorithm. In the k-NN algorithm, the radius $R$ is continuously expanded by a growth factor $\sigma$ until $K$ neighbors are searched. For each local region, the same feature extraction module is for local pattern learning.

\subsection{Deep Feature Extraction}
In this section, we carry out three feature extraction stages to create a higher dimensional point cloud feature representation. The first feature extraction stage is named FE, and the remaining feature extraction stages are named AFE. The FE is an ordinary convolutional structure while the AFE uses spatial and channel 'squeeze and excitation' attention module (scSE) to implement the attention mechanism.

As shown in Fig.~\ref{fig:framework}, we assume that the input of the AFE is point set $\left\{\bm{\vec{x}_1}, \bm{\vec{x}_2}, ..., \bm{\vec{x}_{N_1}}\right\}$ obtained after the FE stage, and we first obtain the cluster centers $\left\{\bm{\vec{x}_{i_1}}, \bm{\vec{x}_{i_2}}, ..., \bm{\vec{x}_{{N_1}'}}\right\}$ with farthest point sampling (FPS). Given the same number of centroids, FPS provides better coverage of the complete point set than random sampling.
 Then we use the ball query approach to obtain the elements of each cluster for feature extraction. Then the constituents of each cluster are obtained by using the Ball Query Method (BQM). The ball query returns $K$ points within a radius of the query point. If the number of points in the radius is less than $K$, the points in the radius are repeatedly sampled until $K$ points are obtained. Here the point with the smallest index in the radius is used as the supplementary point. 

The reason for using different clustering algorithms is that the KNN algorithm has the advantage of ensuring each region contains the same number of points, while the ball-query algorithm ensures that each region has the same size. In the KCE module, we need to capture as much information as possible about the original point cloud (i.e., retain more original points) so KNN algorithm is better than ball-query. However, in the DFE module, we prefer to ensure that each local region has the same perceptual field (i.e., the same region size). It is worth noting that the parameter $R$ of the ball-query algorithm is particularly important in the DFE module. 

\begin{figure}[t!]
\centering{\includegraphics[width=0.48\textwidth]{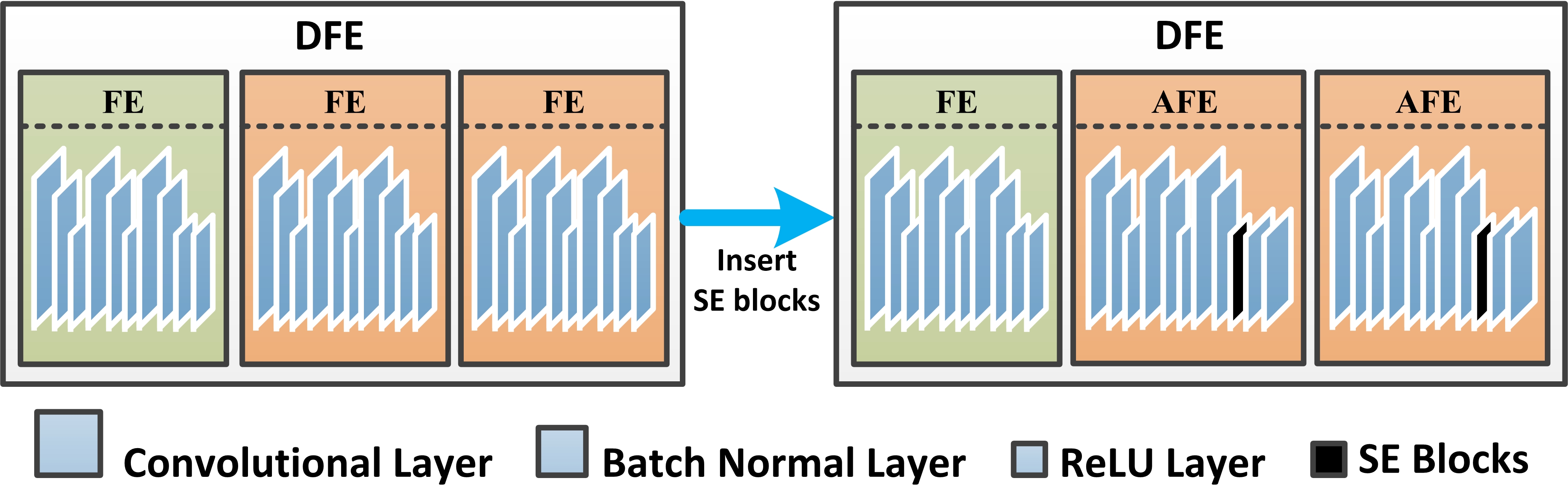}}
\caption{SE Blocks in PKT-PC \label{fig:SE_in_network}}
\end{figure}

Importantly, we integrate the attention mechanism into the feature extraction module of the PointNet backbone network, termed AFE, for feature extraction of cluster2. The internal structure of the AFE is depicted in Fig.~\ref{fig:framework}. To ensure that deeper features of the original point cloud are extracted, this module is repeated twice in the overall structure. According to cognitive science~\cite{yang2020inferring}, humans selectively focus on a portion of all information while ignoring other apparent information due to bottlenecks in information processing. With limited attention resources, this is an excellent way to swiftly filter out high-value information from a vast volume of data. Neural networks incorporating attention mechanisms have made great breakthroughs in various fields, such as image classification ~\cite{mnih2014recurrent}, machine translation~\cite{bahdanau2014neural}~\cite{vaswani2017attention}, and image caption~\cite{xu2015show}. Hu $et.al$~\cite{hu2018squeeze} proposed the Squeezeand-Excitation (SE) block as part of a convolutional neural network (CNN). This block adaptively recalibrates channel-wise feature responses by explicitly modelling interdependencies between channels, achieving excellent results on image classification tasks. Subsequently, Roy, Navab, and Wachinger renamed this SE module spatial squeeze and channel excitation (cSE) module~\cite{roy2018concurrent}, and developed a new channel squeeze and spatial excitation (sSE) module, squeezing channel-wise and exciting spatially. Importantly, they proposed a module termed scSE that merge the two features. We use the scSE module in the proposed network. The location of SE blocks in PKT-PC are shown in Fig.~\ref{fig:SE_in_network}.

\begin{figure}[t!]
\centering{\includegraphics[width=0.48\textwidth]{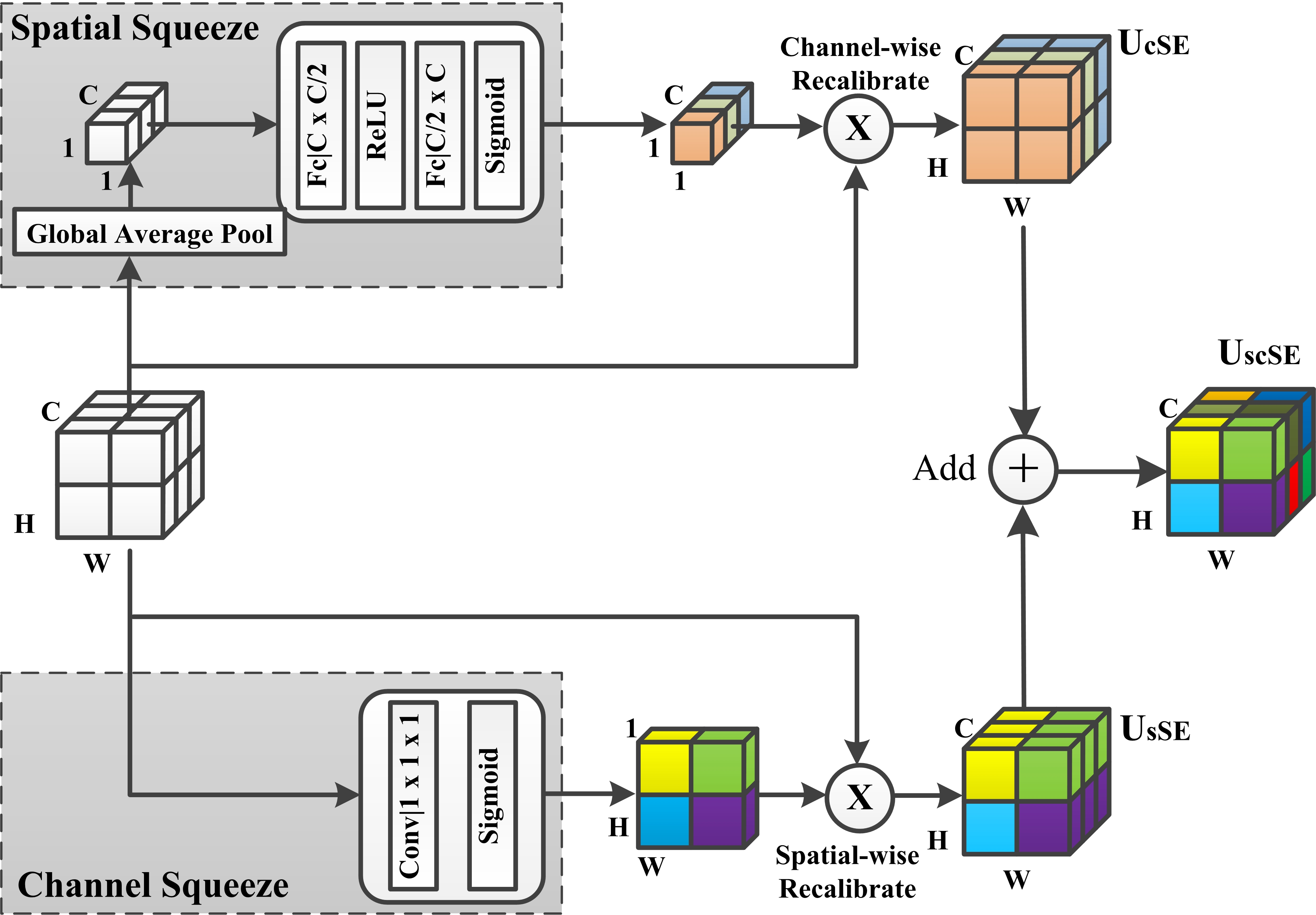}}
\caption{Spatial and Channel 'Squeeze and Excitation' block (scSE). The upper part: Channel Excitation (cSE) module, The lower part: Spatial Excitation (sSE) module. H: the height of image, W: the width of image, C: the channel of image. The blank box represents original input, $U_{scSE}$ represents image after using the self-attention mechanism. \label{fig:scSE}}
\end{figure}

The scSE model consists of the cSE and sSE models, $\mathbf{U_{scSE}}=\mathbf{U_{cSE}}+\mathbf{U_{sSE}}$. The architectural flow is shown in Fig.~\ref{fig:scSE}. The upper part is cSE, while the lower represents sSE. We first introduced the cSE part. Consider the input feature map $ \mathbf{U} = [\bm{\vec{u}_1},\bm{\vec{u}_2},...,\bm{\vec{u}_C}]$, where $\bm{\vec{u}_i} \in \mathbb{R}^{H \times W}$. Spatial squeeze first realized by a global average pooling layer, producing vector $\bm{\vec{z}}  \in \mathbb{R}^{1 \times 1 \times C}$, 
\begin{align}
z_k = \frac{1}{H \times W} \sum_{i}^{H}\sum_{j}^{W}\bm{\vec{u}_k}(i,j).
\label{zi}
\end{align}
Then, encoding the channel-wise dependencies through two fully-connected layers $\bm{\vec{w}}= \mathbf{W_1}(\delta (\mathbf{W_2} \bm{\vec{z}}))$,  with weights $\mathbf{W_1}\in \mathbb{R}^{C \times \frac{C}{2}}$, $\mathbf{W_2} \in \mathbb{R}^{ \frac{C}{2}\times C}$, and ReLU operator $\delta(\cdot)$. Finally, mapping $\bm{w_i}$ into $[0, 1]$ using sigmoid layer $\sigma(\cdot)$. The result $\mathbf{U_{cSE}}$ can be expressed as 
\begin{align}
\mathbf{U_{cSE}} = F_{cSE}(\mathbf{U}) = [\sigma(w_1)\bm{\vec u_1},\sigma(w_2)\bm{\vec u_2},..., \sigma(w_C)\bm{\vec u_C}].
\label{UcSE}
\end{align}

As the network learns, the channel weight $\sigma(\vec{w}_1)$ are dynamically adjusted, turning that the less important channels are ignored and the important ones are emphasized. Unlike the cSE module, the sSE module cares more about spacial relationships. Thus, the input tensor $\mathbf{U}$ can be described as $[\bm{\vec{u}^{1,1}},\bm{\vec{u}^{1,2}},...,\bm{\vec{u}^{i,j}},...,\bm{\vec{u}^{H,W}}]$, where $\bm{\vec{u}^{i,j}} \in \mathbb{R}^{ 1\times 1 \times C}$, and $(i,j)$ ($i\in\left\{ 1,2,...,H\right\}$, $j\in\left\{ 1,2,...,W\right\}$) represents its corresponding spatial location. The channel squeeze operation is achieved through a convolution layer, $\bm{\vec{w}}=\mathbf{U} \times \mathbf{W_{sq}}$ with $ \mathbf{W_{sq}} \in \mathbb{R}^{ 1\times 1 \times C\times 1}$. This operation generates a projection tensor $\bm{\vec{w}}\in \mathbb{R}^{ H\times W}$, where $w_{i,j}$ is the linear combination of all channels in spatial location $(i,j)$. Same as cSE, $w_{i,j}$ needs to be re-scaled into $[0,1]$ with a sigmoid layer $\sigma(\cdot)$ to be a spatial weight. Finally, the spatial recalibration of input $\mathbf{U}$ is
\begin{equation}
\begin{split}
\mathbf{U_{sSE}} &= F_{sSE}(\mathbf{U})\\
&= [\sigma(w_{1,1})\bm{\vec u^{1,1}},...,\sigma(w_{i,j})\bm{\vec u^{i,j}},..., \sigma(w_{H,W})\bm{\vec u^{H,W}}],
\end{split}
\label{UsSE}
\end{equation}
where activation $\sigma(w_{i,j})$ indicates the relative importance of location $(i,j)$ for a given feature map, enabling more attention to some relevant spatial locations while ignoring less important space. By combining cSE and sSE, scSE provides more importance of the point which are relevent both spatially and channel-wise, that encourages the network to learn more meaningful feature maps.

\subsection{Progressive Prediction Mechanism}
\begin{figure}[t!]
\centering{\includegraphics[width=0.4\textwidth]{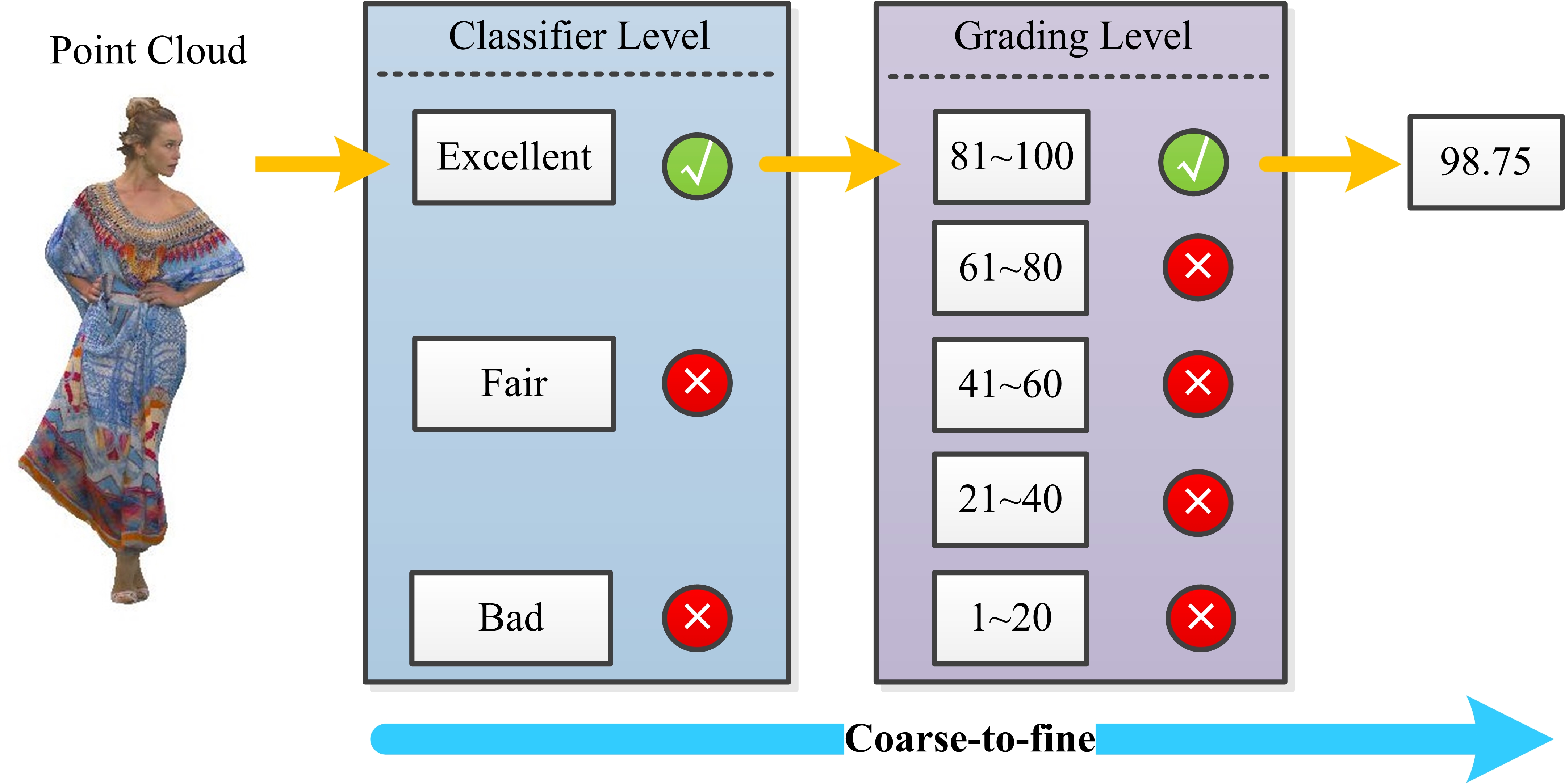}}
\caption{Illustration of the human scoring process from coarse to fine} \label{fig:knowledge_structure}
\end{figure}

Referring to the process of evaluating the quality of point clouds by human perceptual system, we can find that there exists a native hierarchical dependency. That means that humans always first determine a point cloud coarse-grained quality, i.e., excellent, fair, or bad, and then, under the branch of the quality type, give a specific quality score by relying
on subtle visual cues. Fig.~\ref{fig:knowledge_structure} illustrates the coarse-to-fine principle.

In the proposed network, we adopt a two-step strategy to train the PKT-PCQA model: 1) quality level classification; 2) quality score prediction. In step 2), the feature extraction part of the PKT-PCQA model is initialized with related part in step 1), then fine-tune the quality prediction network on the first quality classification basis. In the process from coarse to fine, there are two key parts, namely, the classification of quality level and the training of the network.

(1) Quality Level Classification
\begin{figure*}[t!]
\centering{\includegraphics[width=1\textwidth]{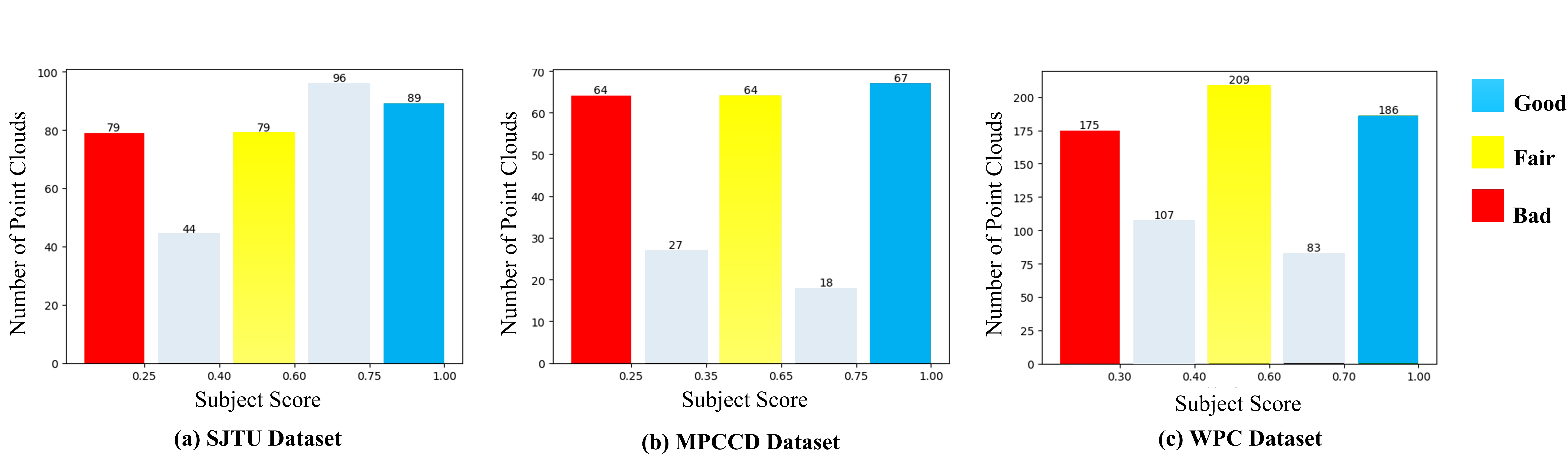}}
\caption{Quality Level Division in Different Datasets\label{fig:score2level}}
\end{figure*}

The classification of quality levels is based on the statistical properties of quality scores in the dataset. That is, within a certain range, the number of point clouds at each level is guaranteed to be approximately equal in each dataset. The specific classification will be discussed in the experimental settings section. In this paper, we only use three labels: excellent, fair and bad, in order to maximize the distinction of different labels.

(2) Model Training

In the classification task, the parameters $\mathbf{W_{fc}}$ of the feature extraction part and $\mathbf{W_{c}}$ of the classification part are determined by minimizing the loss function $cross-entropy$
\begin{align}
(\mathbf{\hat{W}_{fc}},\mathbf{\hat{W}_{c}})=\arg \min L_1(\bm{\vec{x_i}}[y_i];\mathbf{W_{fc}},\mathbf{W_{c}}),
\label{class-loss-function}
\end{align}
\begin{align}
L_1(\bm{\vec{x_i}},y_i)=-log\left(\frac{exp(\bm{\vec{x_i}}[y_i])}{\begin{matrix}\sum_{j=1}^{c} exp(\bm{\vec{x_i}}[j])\end{matrix}}\right),
\label{CrossEntropy}
\end{align}
where $\bm{\vec{x_i}}$ represents a specific prediction value vector for point cloud $i$, $y_i$ is the label value used as an index, and $c$ is the number of types of labels.

For MOS prediction network, the model parameters of the feature extraction part are initialized as $\mathbf{\hat{W}_{fc}}$, while the prediction unit is randomly initialized, represented as $\mathbf{W_p}$. In prediction task, we care more about how well the prediction curve fits the original score curve, rather than the difference between individual scores \cite{mantiuk2012comparison}. Therefore, the related model parameters are fine-tuned through minimizing the loss function measures the difference in fit. The parameters of the model are determined by (\ref{k-batch-train}) after training the k-th batch.
\begin{align}
(\mathbf{\hat{W}_{fp}},\mathbf{\hat{W}_{p}})&=\arg \min L_2(\bm{\vec{x_k}};\mathbf{\hat{W}_{fc}},\mathbf{W_{p}}),
\label{k-batch-train}\\
L_2(\bm{\vec{x_k}},\bm{\vec{y_k}})&=(1-P(\bm{\vec{x_k}},\bm{\vec{y_k}}))^2,\label{pred-loss-function}\\
P(\bm{\vec{x_k}},\bm{\vec{y_k}})&=\frac{E[(\bm{\vec{x_k}}-\mu_x)(\bm{\vec{y_k}}-\mu_y)]}{\sigma_x\sigma_y}\notag \\
&=\frac{E[(\bm{\vec{x_k}}-\mu_x)(\bm{\vec{y_k}}-\mu_y)]}{\sqrt{\begin{matrix}\sum_{i=1}^{k}(x_i-\mu_x)^2\end{matrix}\begin{matrix}\sum_{i=1}^{k}(y_i-\mu_y)^2\end{matrix}}},
\label{pearson-correlation}
\end{align}
where $P(\bm{\vec{x_k}},\bm{\vec{y_k}})$ is pearson linear correlation coefficient (PLCC) with the range is [-1,1], and "1" means positive correlation, "-1" means negative correlation, "0" means uncorrelated. In the quality prediction task, our goal is to maximize $P(\bm{\vec{x_k}},\bm{\vec{y_k}})$. Therefore, we can obtain the parameters of the quality prediction network after mining the loss function (i.e., L2).

\section{Experimental Results and Discussion}
\label{sec:Experimental Results and Discussion}
In this section, we describe the point cloud datasets and assess the performance of the proposed method. 
\subsection{Experimental Settings}
PKT-PCQA was implemented by using PyTorch on a computer with a 3.5 GHz CPU and GTX 1080Ti GPU.
  
(1) Dataset
  
The proposed method was evaluated on three 3D point cloud: SJTU-PCQA ~\cite{yang2020predicting}, M-PCCD~\cite{alexiou2019comprehensive}, and the Waterloo Point Cloud (WPC)~\cite{su2019perceptual}~\cite{liu2022perceptual}. The quality level division on each dataset is shown in Fig.~\ref{fig:score2level}.
\begin{itemize}
\item  {\textbf{SJTU} consists of 9 reference point clouds (\emph{redandblack, loot, longdress, hhi, soldier, ricardo, ULB Unicorn, Romanoillamp, statue, shiva}) and 378 distorted versions (42 distorted items per reference item through 6 degradation types: OT: Octree-based compression, CN: Color Noise, DS: Downscaling, D+C: Downscaling and Color noise, D+G: Down-scaling and Geometry Gaussian noise, GGN: Geometry Gaussian noise and, C+G: Color noise and Geometry Gaussian noise).}

\item  {\textbf{M-PCCD} consists of 8 reference point clouds (\emph{loot, longdress, soldier, Romanoillamp, amphoriskos, biplane, head, the20smaria}) and 232 distorted versions (29 distorted items per reference item through 5 types of compression G-PCC with Octree coding and Lifting coding, G-PCC with Octree coding and RAHT coding, G-PCC with Triangle soup coding and Lifting coding, G-PCC with Triangle soup coding and RAHT coding, V-PCC).}

\item  {\textbf{WPC} consists of 20 reference point clouds (\emph{bag, banana, biscuits, cake, cauliflower, flowerpot, glasses case, honeydew melon, house, litchi, mushroom, pen container, pineapple, ping-pong bat, puer tea, pumpkin, ship, statue, stone, tool box}) and 740 distorted versions 37 distorted items per reference item through five degradation types TMC1: G-PCC with Triangle soup coding, TMC2: G-PCC with Octree coding, TMC3:V-PCC, Noise: White Gaussian noise, Downsample: Octree-Based Downsampling).}
\end{itemize}

(2) Parameter Setting  
  
The following parameter values were used
\begin{figure}[t!]
\centering{\includegraphics[width=0.4\textwidth,height=0.18\textheight]{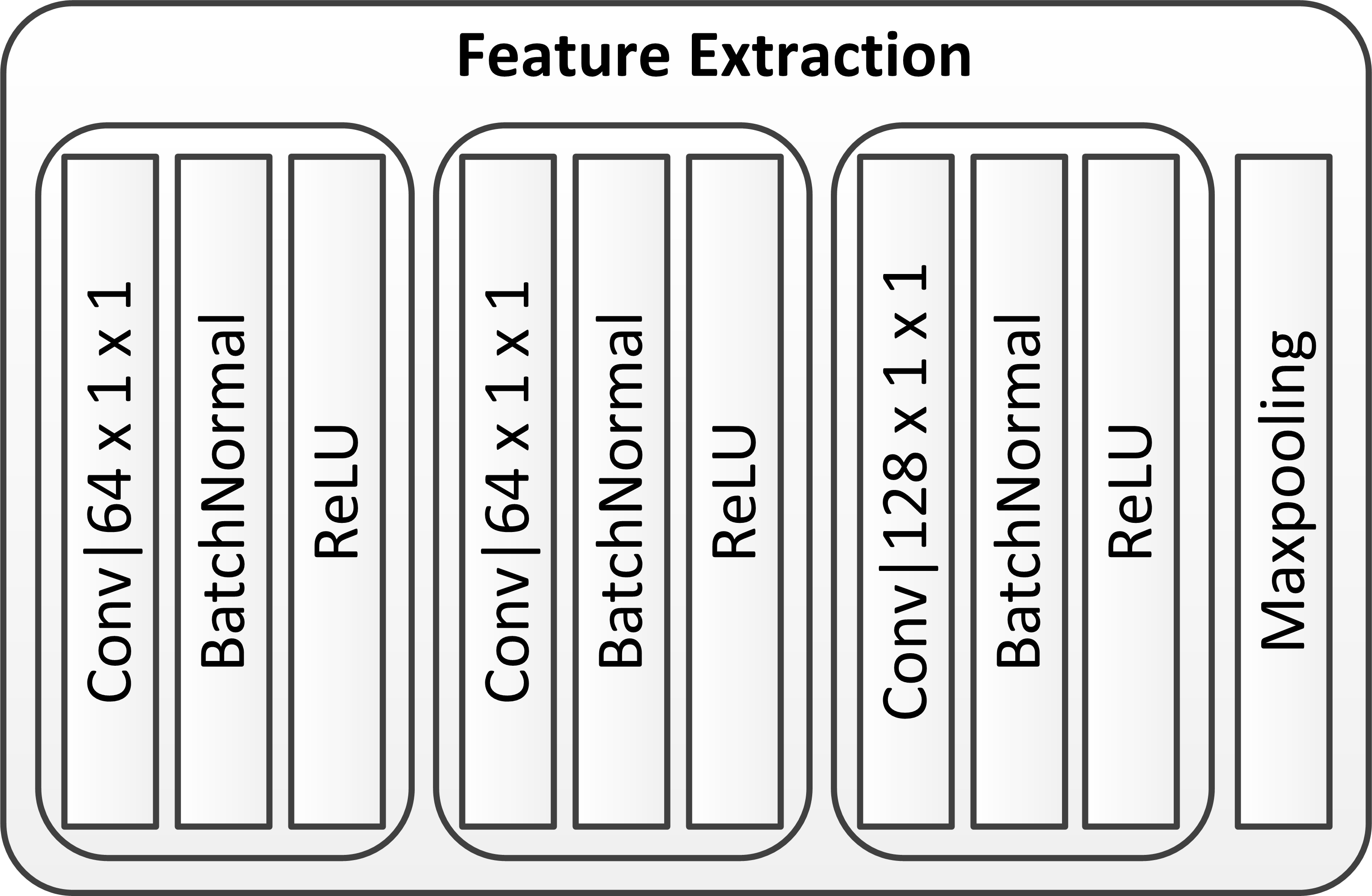}}
\caption{Feature Extraction Unit in DFE \label{fig:AFE1}}
\end{figure}
\begin{figure*}[t!]
\centering{\includegraphics[width=1\textwidth]{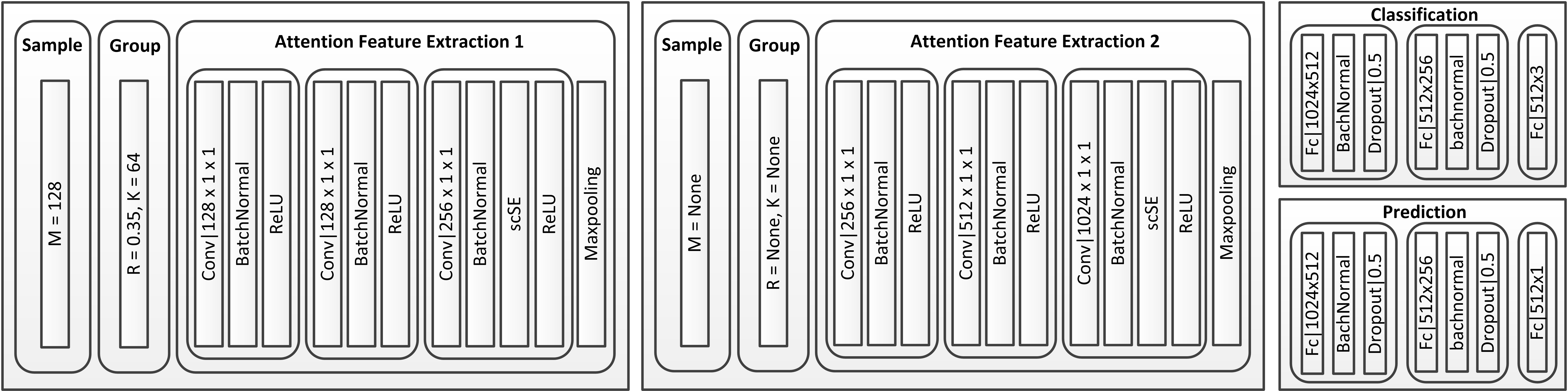}}
\caption{Deep feature extraction, classification and prediction\label{fig:Deep Feature Extraction}}
\end{figure*}

\begin{itemize}
\item  {\textbf{$\beta,L,K$ in Key Clusters Extraction} \\
$\beta=1024,L=4$ in key-point extraction, and $K=16$ in clustering. These values of parameters $\beta,K$ can be proved in Ablation Test, while $L$ follows the values suggested in the reference paper \cite{yang2020inferring}. The parameter settings of the FE units are shown in Fig.\ref{fig:AFE1}.}

\item  {\textbf{$R$ in Deep Feature Extraction} \\
DFE includes sampling, grouping and attention feature extraction twice, and the specific structure and parameters are shown in Fig.\ref{fig:Deep Feature Extraction}, where scSE implements the attention function of the model. The parameter $R$ determines the size of the perceptual field. It is set to $0.4$ in the classification task and $0.35$ in the prediction task.}

\item  {\textbf{Progressive Prediction Mechanism}\\
The Classification and Prediction branches have a similar architecture; the only difference between them is the last layer. In the classification task, the out-channel of last layer in classification task is the number of quality level, while that of prediction task is one.

In the classification task, the evaluation criterion is $CrossEntropy$ loss function, while in the prediction quality branch, we considered the PLCC (\ref{pearson-correlation}) and Spearman's rank-order correlation coefficient (SROCC) computed as 
\begin{align}
SROCC=1-\frac{6 \sum_{i} d_{i}^{2}}{I\left(I^{2}-1\right)},
\label{SRCC}
\end{align}
where $I$ is the number of the tested point clouds and $d_{i}$ is the rank difference between the ground truth $MOS$ and the predicted $MOS$ of the $i$-th point cloud.}

\item  {\textbf{Parameters in Multistep Training Mechanism}\\
We used the Adam optimization technique with default parameters to train both the classification and prediction tasks. The batch size was set to 16 in training processing and 32 in testing processing, with a total epoch number of 200. In the classification task, the learning rate was initialized to 0.001 and subsequently reduced using $stepLR$, a strategy for adjusting the learning rate at equal intervals, with a step size of 20, and the learning rate adjustment multiplier $\gamma = 0.7$. In the prediction task, the learning rate was set to 0.0001, using $stepLR$ was used as a reduction strategy with step size 30, and learning rate adjustment multiplier $\gamma = 0.7$.}
\end{itemize}

\subsection{Ablation Test}
To fully evaluate our proposed model, we conduct the following experiments.

(1) Key Clusters Extraction\\
The Key Clusters Extraction (KCE) aims at simplifying the point cloud without losing critical information . Common reduction strategies include point cloud sampling and point cloud segmentation, which correspond to the global feature and local feature of a point cloud. However, in point cloud quality prediction task, neither of them is enough to represent the original point cloud. Extracting key clusters of point cloud needs to take into account global and local information. To validate this hypothesis, we evaluate point cloud extraction strategies using only global information, only local information, and both global and local information. Before making the above comparison, we need to choose a representative algorithm for each strategy.

\begin{figure}[t!]
\centering{\includegraphics[width=0.43\textwidth]{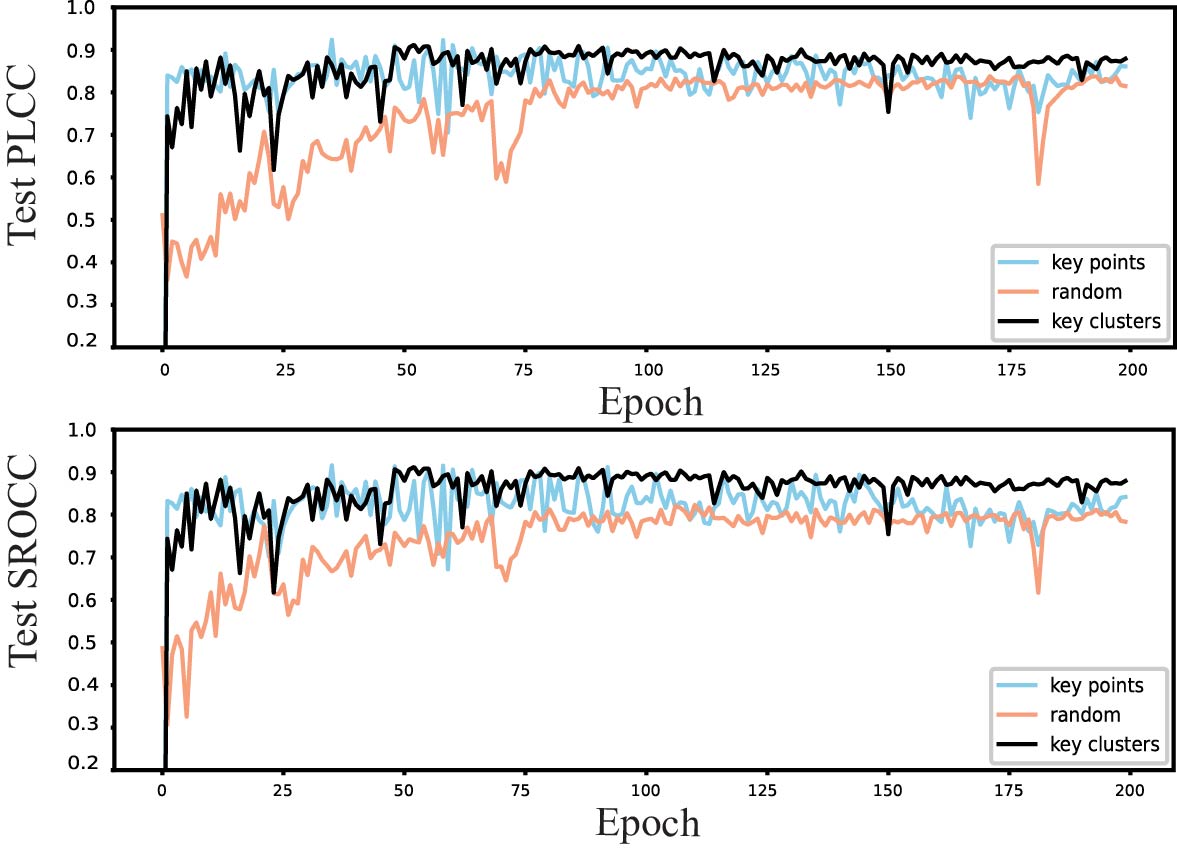}}
\caption{Comparison between farthest point sampling (fps), random sampling, and key point extraction (reference). The PLCC and SROCC as computed for \label{fig:global feature extraction strategy}}
\end{figure}

We first compared several global information extraction strategies, including random sampling, farthest point sampling, and key-point extraction. The result of using them separately to complete the prediction of point cloud quality is shown in Fig.\ref{fig:global feature extraction strategy}. Using key points to represent the global feature gives the best result. Therefore, extracting key points method is used as a proxy for global information extraction strategy to participate in the comparison with the other two feature extraction strategies.

A popular technique for representing point clouds using local information is to partition them into chunks. However, in the task of predicting the quality score of a point cloud, it is challenging to determine the score of the entire point cloud using this method because each chunk receives a score, and it is inappropriate to choose one of them to represent the whole or determine the average of all chunks. Here, we only contrast the effects of the two strategies on the outcomes that is key points and key clusters. We can see that using key clusters gives better result than just using key points.

\begin{figure}[t!]
\centering{\includegraphics[width=0.45\textwidth]{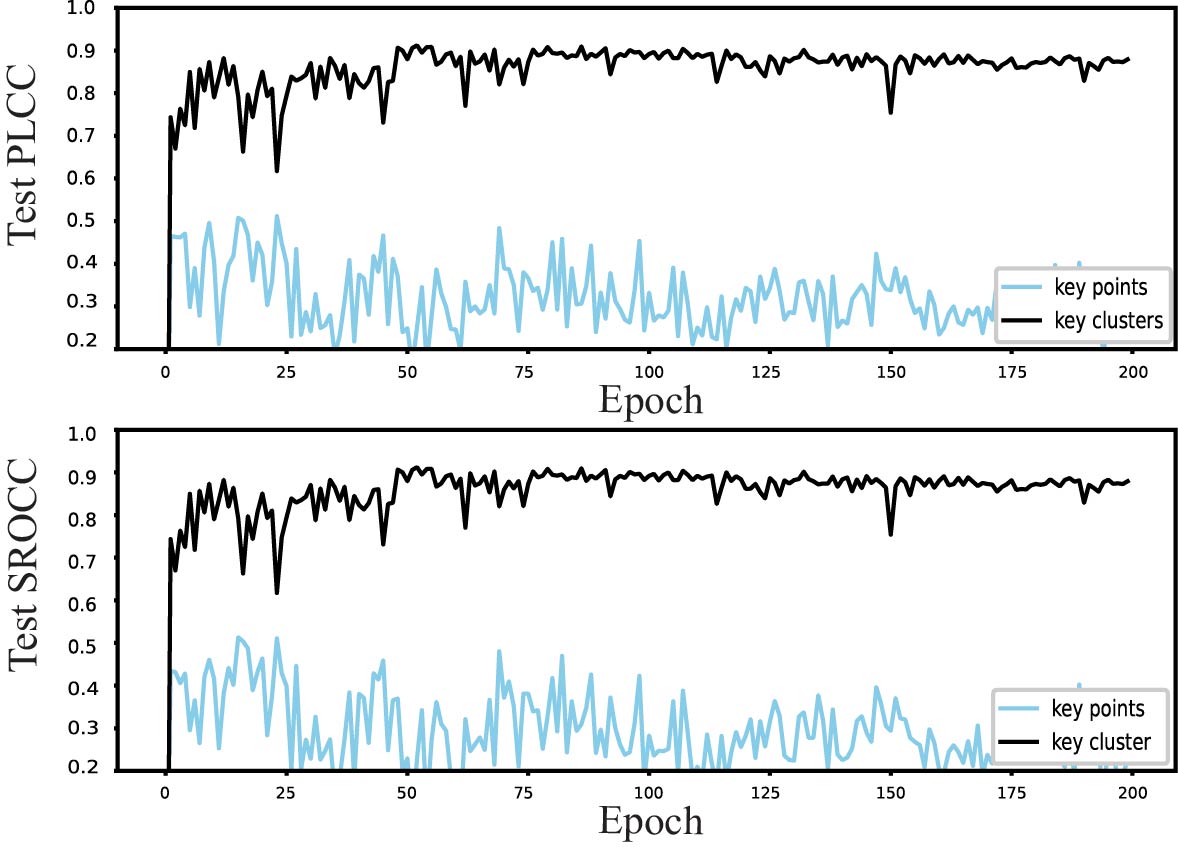}}
\caption{Comparison between keys points and key clusters \label{fig:feature extraction strategy}}
\end{figure}

Comparing the influence of these three reconstruction methods in point cloud prediction task, as shown in Fig.\ref{fig:feature extraction strategy}, the proposed method that considers both global and local features is a better strategy to extract the key clusters of point cloud. The results in Fig.\ref{fig:feature extraction strategy} also proved that strategy can provide more efficient information to subsequent network in point cloud quality evaluation task.

To find the most suitable number of the new point cloud, we also test several combinations of keypoints and local points $\beta \times K$ (i.e. $1024 \times 16, 1024 \times 32, 2048 \times 16, 512 \times 16$).
  
The network structure is shown in Fig.\ref{fig:AFE1} and Fig.\ref{fig:global feature extraction strategy} with the same training strategies. The experimental results are shown in Table~\ref{different number of points}. We can see that the best combination of the number of key point and local point number is $1024 \times 16$. The reason is that too many key points may cause information redundancy in the original point cloud, affecting the information provided by valid points, while too few key points cannot provide enough information, so the number of key points needs to be apposite.
\begin{table}[t!]
\centering \caption{Different Combinations of Global and Local Points}
\begin{tabular}{cccc}
\toprule \hline $\beta$   &K    &PLCC     &SROCC\\\hline
\rowcolor{mygray} 512  &16     &0.8803         &0.8718         \\
 \textbf{1024}  &\textbf{16}     &\textbf{0.9122}         &\textbf{0.9316}         \\
\rowcolor{mygray} 2048  &16     &0.8448         &0.8485          \\
1024  &32     &0.8773         &0.8991                \\\hline \bottomrule
\end{tabular}
\label{different number of points}
\end{table}

\begin{table}[t!]
\centering \caption{Comparison for Different R}
\begin{tabular}{ccc}
\toprule \hline R    &PLCC     &SROCC\\\hline
\rowcolor{mygray} 0.4     &0.8892         &0.9049         \\
 \textbf{0.35}     &\textbf{0.9122}         &\textbf{0.9316}         \\
\rowcolor{mygray} 0.3     &0.8865         &0.8993          \\\hline \bottomrule
\end{tabular}
\label{Comparison among Different R}
\end{table}

(2) Deep Feature Extraction

The radius $R$, which represents the perceptual field size of the depth extraction network is a critical parameter. If $R$ is too large, the  tightness of neighbors and clustering centers cannot be guaranteed because the range of random sampling is too large. On the other hand, if $R$ is too small, not enough neighbor points can be collected, leading to the reduction of effective information obtained in the local area. We used $1024 \times 16$ point cloud key clusters to test the network with radius 0.4, 0.35, and 0.30. The results in Table~\ref{Comparison among Different R} show that $R=0.35$ provides the maximum prediction accuracy.

(3) Attention Mechanism\\
The impact of the attention mechanism on the model depends on the type of SE block and its location. To fully explore it, we compare the effects of the three type SE blocks at different locations in the model. As for the location, we discuss the different locations of SE in KCE and DFE modules separately.

\begin{figure}[t!]
\centering{\includegraphics[width=0.45\textwidth,height=0.35\textheight]{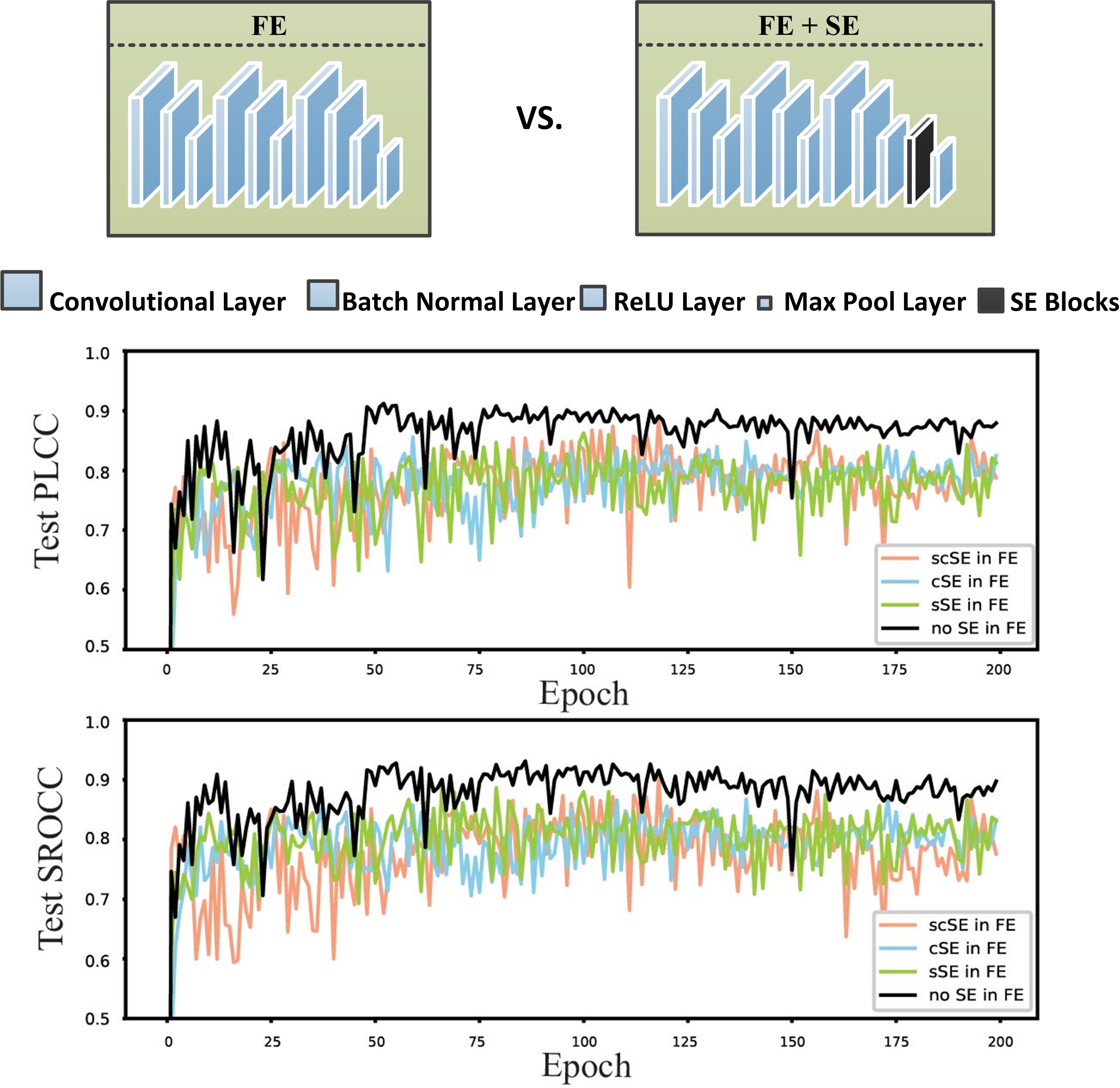}}
\caption{Attention mechanism for initial feature map generation, where black line represents no SE module in FE unit of DFE, and red, blue and green one are scSE, cSE, and sSE modules respectively. \label{fig:SE in KPE}}
\end{figure}

\begin{figure*}[htbp]
    \centering
    \subfigure[Optimal SE Module]{
    \label{a}
    \includegraphics[width=0.9\textwidth, height=4cm]{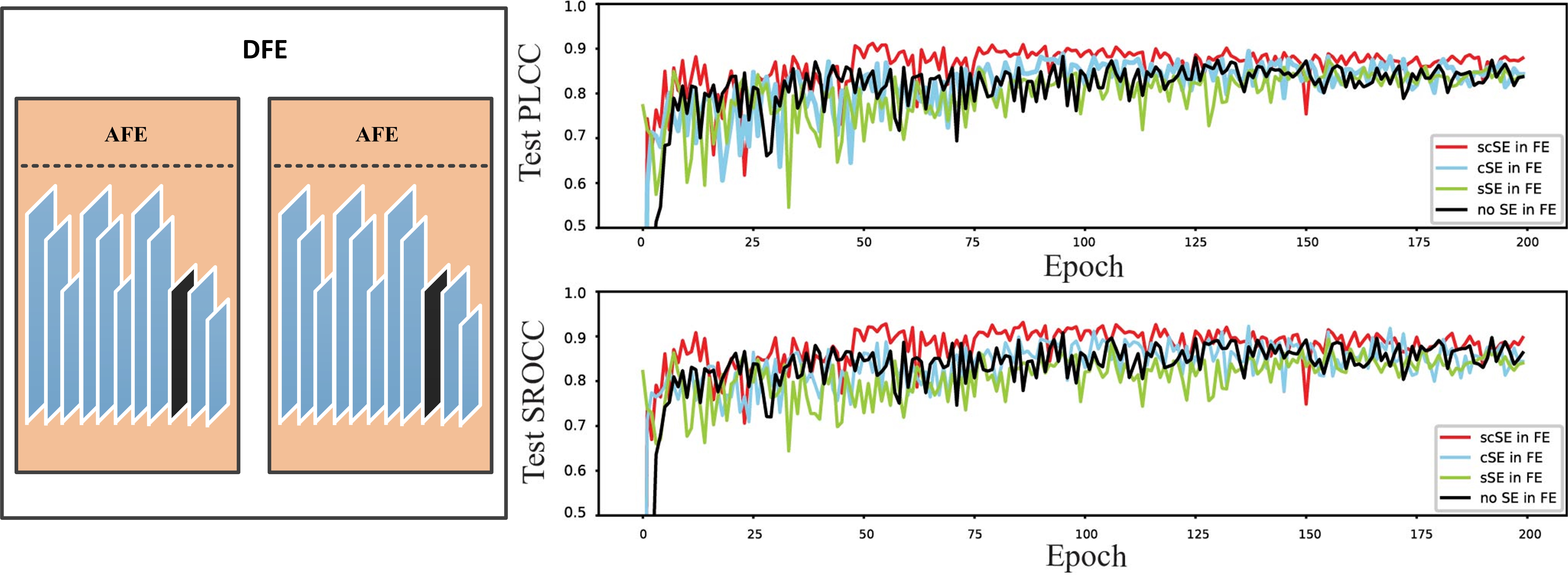} 
    }
\subfigure[SE block after ReLU ]{
\label{b}
\includegraphics[width=0.9\textwidth, height=4cm]{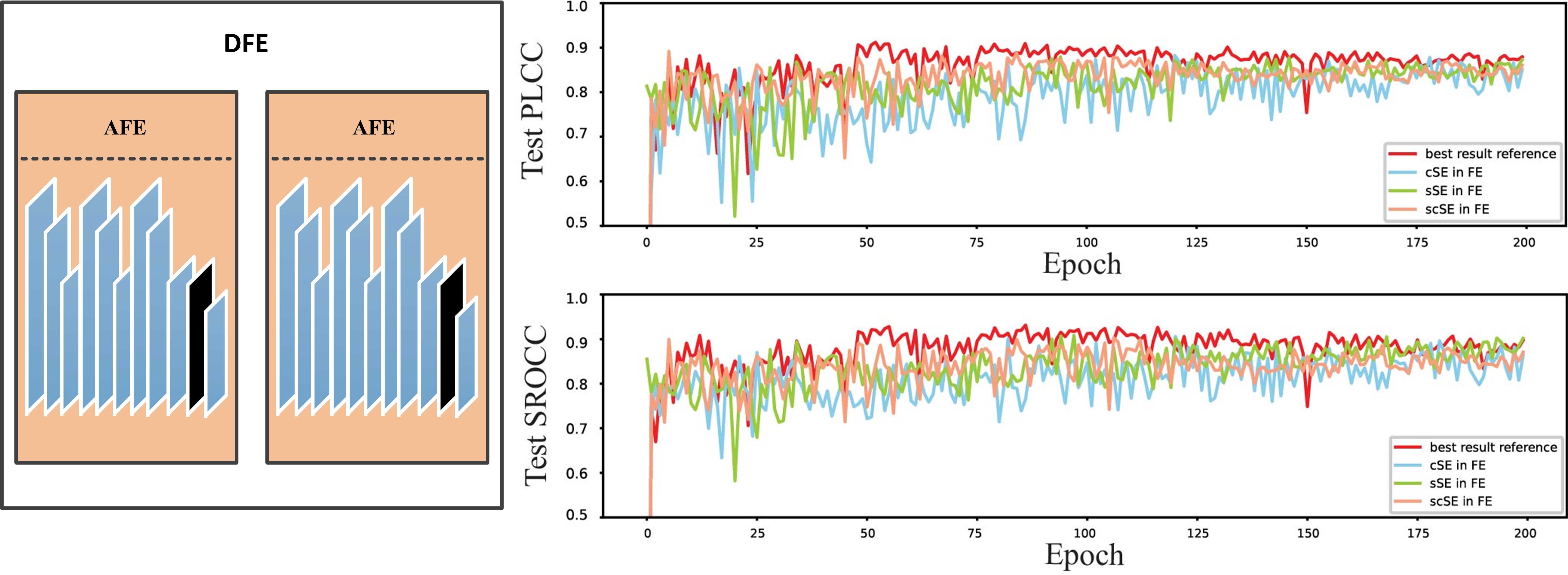} 
}
\subfigure[Only First AFE has SE Block]{
\label{c}
\includegraphics[width=0.9\textwidth, height=4cm]{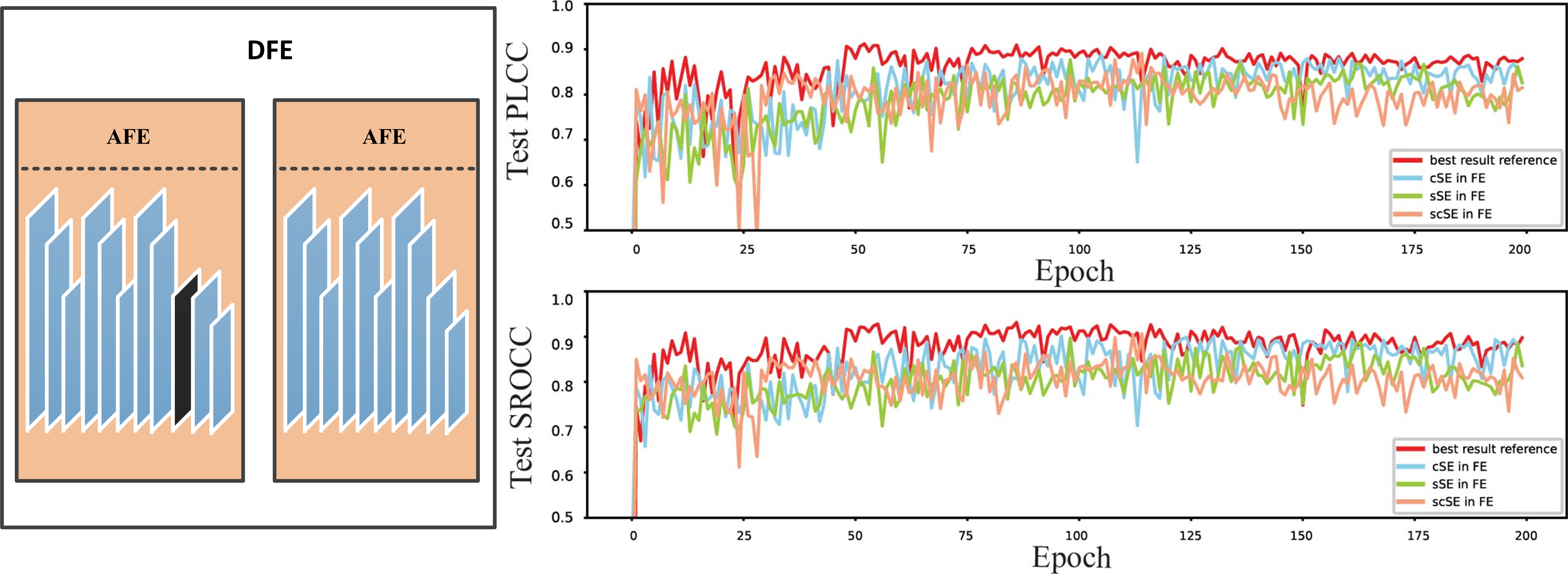} 
}
\subfigure[Only Second AFE has SE Block]{
\label{d}
\includegraphics[width=0.9\textwidth, height=4cm]{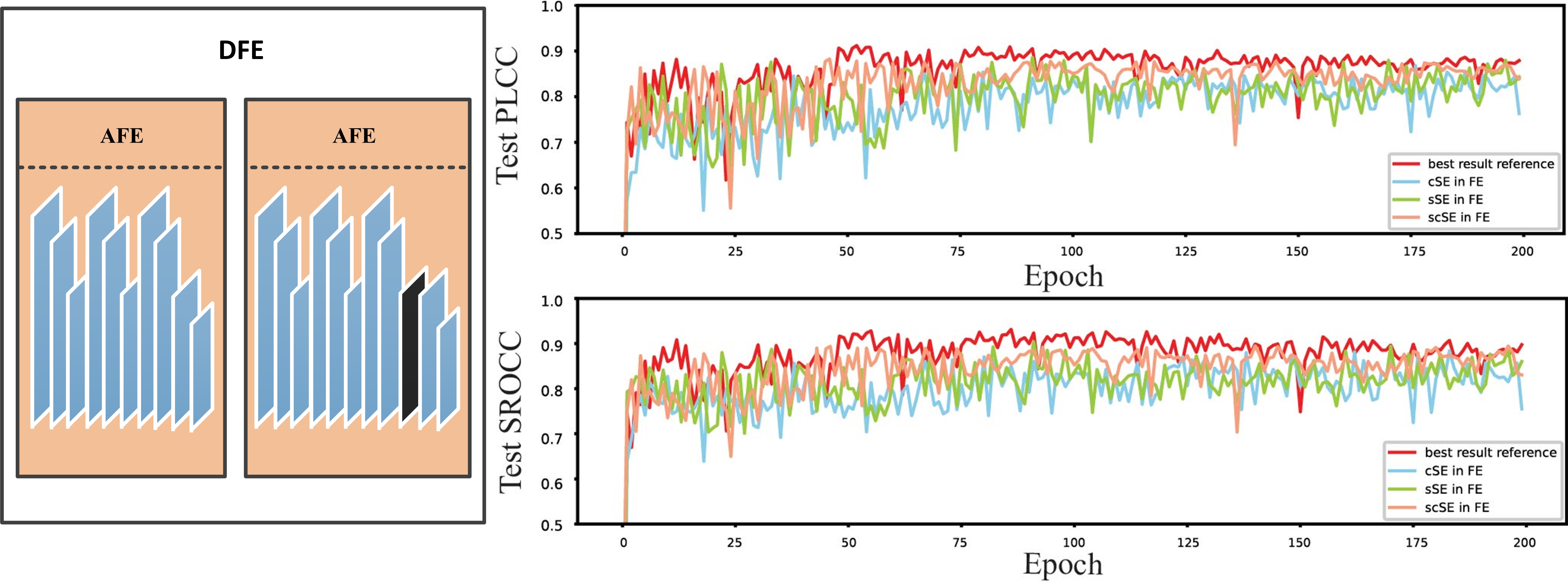} 
}
\subfigure[Two SE Blocks in One AFE]{
\label{e}
\includegraphics[width=0.9\textwidth, height=4cm]{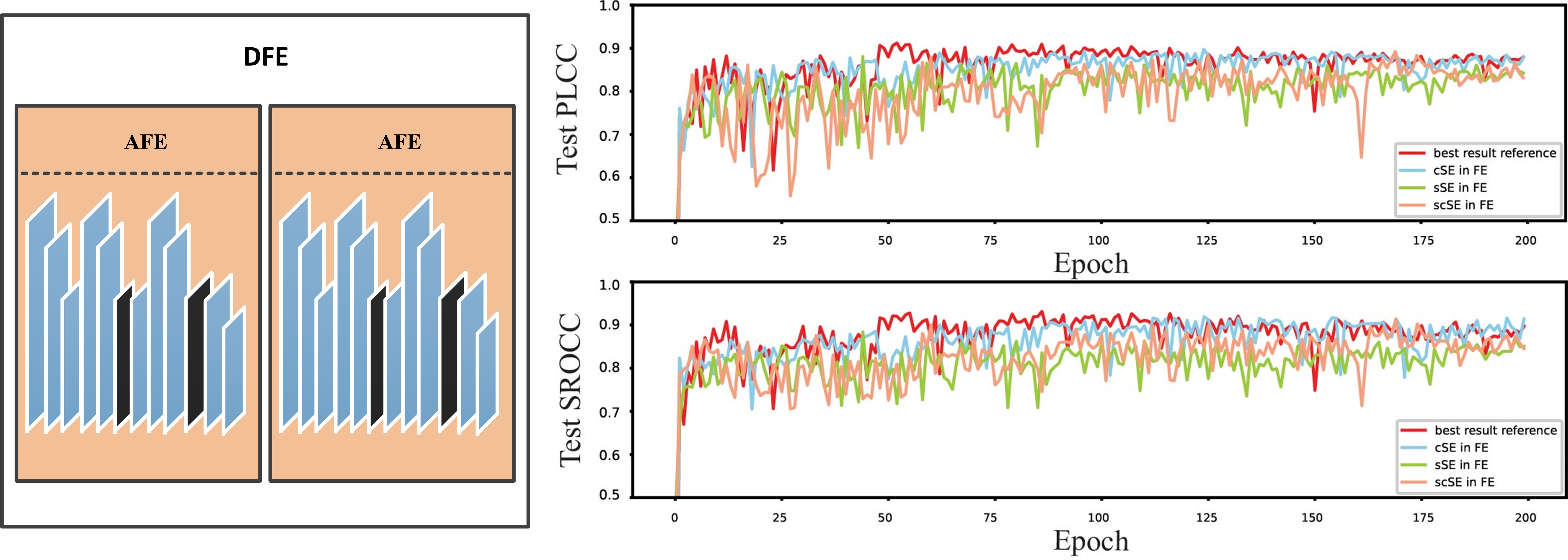} 
}
\subfigure[\empty]{
\includegraphics[width=0.8\textwidth, height=0.5cm]{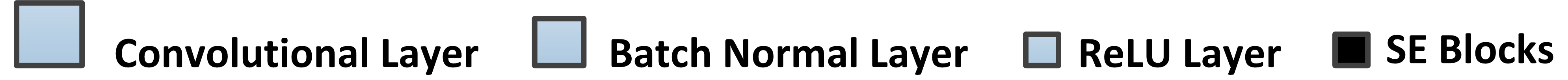}
}
\caption{Different Attention Mechanism in Different Place}
\label{fig:differentattention}
\end{figure*}

\begin{table*}[t!]
\centering \caption{Attention Mechanism Comparison in PKT-PCQA}
\begin{tabular}{c|c|c|c|c|c|c}
\toprule \hline 
\multicolumn{1}{l|}{}& \multicolumn{3}{c|}{PLCC}
&\multicolumn{3}{c}{SROCC}\\
\cline{2-7}
\multicolumn{1}{l|}{}&cSE&sSE&scSE&cSE&sSE&scSE\\
\hline
\rowcolor{mygray} \multicolumn{1}{l|}{(a) Optimal Model}&0.8946&0.8742&\textbf{0.9122}&0.9227&0.8944&\textbf{0.9316}\\
\hline
\multicolumn{1}{l|}{(b) SE block after ReLU}&0.8827&0.8797&0.8915&0.9065&0.9083&0.9032\\
\hline  
\rowcolor{mygray} \multicolumn{1}{l|}{(c) Only First AFE has SE Block}&0.8880&0.8768&0.8916&0.9020&0.8969&0.9065\\
\hline
\multicolumn{1}{l|}{(d) Only Second AFE has SE Block}&0.8683&0.8869&0.8830&0.8845&0.9044&0.8952\\
\hline  
\rowcolor{mygray} \multicolumn{1}{l|}{(e) Two SE Blocks in One AFE}&0.8976&0.8926&0.8808&0.9197&0.8839&0.9136\\
\hline
\bottomrule
\end{tabular}
\label{AT-comparison}
\end{table*}
\begin{itemize}
\item  {Attention Mechanism for Initial Feature Map Generation}
Generally speaking, adding the attention mechanism can improve the performance of the proposed network. However, in our proposed network, we extract key points in KCE module, which may lead to the failure of the spatial attention mechanism in initial feature map generation. Also, since there are only six channels for each point in this step which represents their three-dimensional coordinates and color information, it is difficult to say which channel is more important. Therefore, the channel attention mechanism may not have much effect in initial feature map.\\
To check this assumption, we conducted comparative experiments on a key clusters ($1024 \times 16$) . SE Blocks in the DFE part are set as scSE. The result is shown in Fig.\ref{fig:SE in KPE}. It is obviously that without any SE module in initial feature map generation can get better prediction result.

\begin{table}[t!]
\centering 
\caption{Overall Comparative Experiment on Attention Mechanism}
\begin{tabular}{c|cc|cc}
\toprule \hline
$\beta \times K$                &KCE                      &DFE              &PLCC               &SROCC \\\hline
\multirow{3}{*}{1024 $\times$ 16} &\cellcolor{mygray}{-}     &\cellcolor{mygray}{-}              &\cellcolor{mygray}{0.8827}          &\cellcolor{mygray}{0.9077} \\
                                &\textbf{-}               &\textbf{scSE}    &\textbf{0.9122}  &\textbf{0.9316}\\
                             &\cellcolor{mygray}  {scSE}  &\cellcolor{mygray}{scSE}   &\cellcolor{mygray}{0.8869}           &\cellcolor{mygray}{0.9047}\\\hline
 \bottomrule
\end{tabular}
\label{total-AT-comparison}
\end{table}
\item {Attention Mechanism in Deep Feature Extraction}\\
Different with the initial feature map, the deep feature of point cloud is much more abstract, that is why attention mechanism can greatly improve the performance of the original model. The degree of improvement depends not only on the type of the attention unit, but also on its location.

Fig.\ref{fig:differentattention} shows the effect of different attention mechanism types at different locations of the AFE module. In the experiments, initial feature extraction did not include an adopt SE block. The best prediction accuracy of each model presented in Table \ref{AT-comparison}.

Fig.~\ref{fig:differentattention}.\subref{a} shows the performance of different SE modules at the ideal position, and the black line result means that no any SE module is added, and the red, blue and green line results represent the performance of adding scSE, cSE, and sSE, respectively. We can see that the scSE module provides the best performance. We also tested the performance of different SE modules in various locations, as shown in Fig.~\ref{fig:differentattention}.\subref{b} to Fig.~\ref{fig:differentattention}.\subref{e}. Fig.~\ref{fig:differentattention}.\subref{b} allows us to evaluate whether the SE block should be before or after the ReLU layer. Fig.~\ref{fig:differentattention}.\subref{c} and Fig.~\ref{fig:differentattention}.\subref{d} can determine if the SE block is appropriate in only one feature extraction step of the DFE module. Besides, Fig.~\ref{fig:differentattention}.\subref{e} helps to decide whether more than one SE block should be added to an AFE module. We can see that adding scSE to the last convolutional layer of each AFE module gives the best results.
\end{itemize}

Finally, we used $1024 \times 16$ key clusters of the point cloud to examine several  combinations of the attention module in the KCE and DFE modules (Table \ref{total-AT-comparison}). The attention module is set as scSE module since prior experiments have shown that this module can significantly improve model performance. We can observe that using the scSE module only in the DFE part leads to the highest score prediction accuracy.

(4) Progressive Prediction Mechanism

\begin{figure}[t!]
\centering{\includegraphics[width=0.49\textwidth]{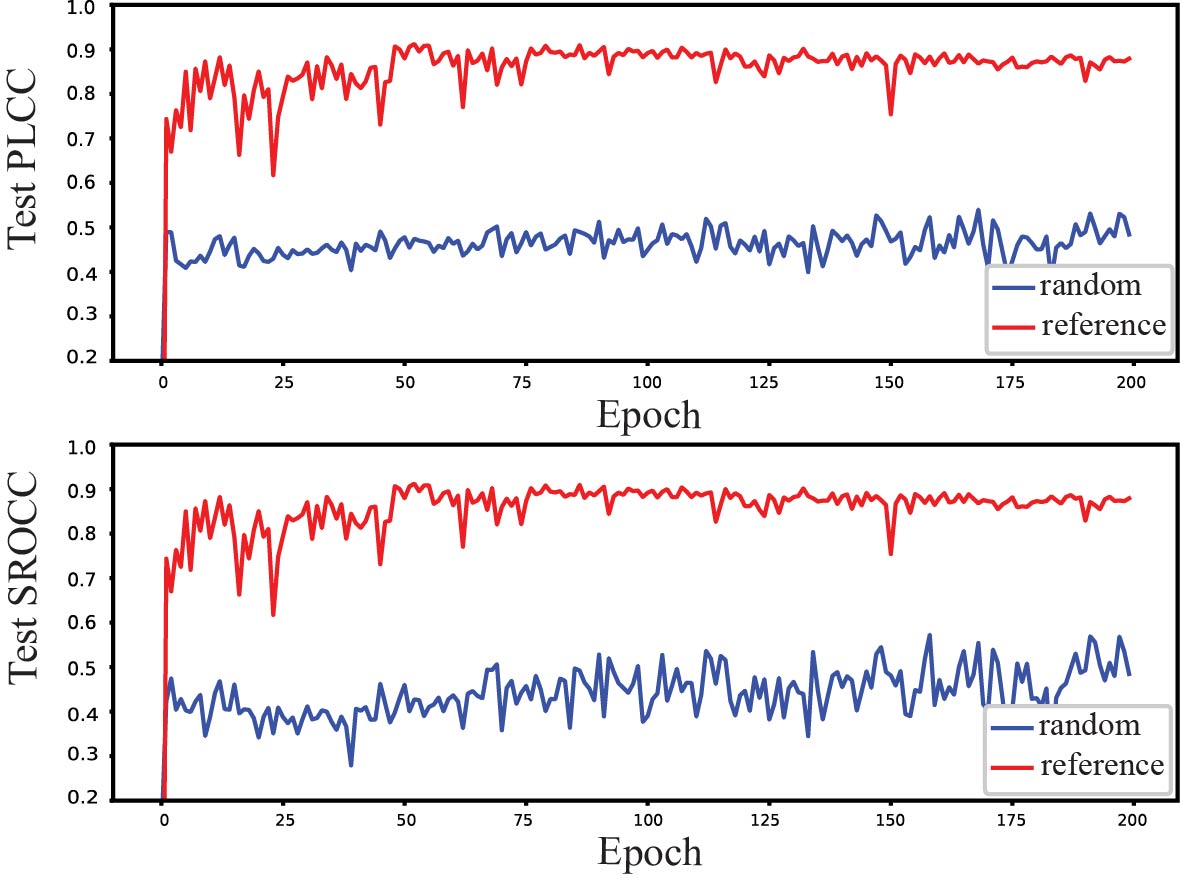}}
\caption{Coarse-to-Fine Strategy and Random Initialization  \label{fig:coarse2fine-comparison}}
\end{figure}

The proposed network predicts the quality of the tested point cloud using a coarse-to-fine progressive mechanism from quality classification to quality prediction. To verify the importance of quality classification information initialization for the quality prediction task, we compared with random initialization. To make a fair comparison,
all the parameters of the prediction task were identical. The results are shown in Fig.\ref{fig:coarse2fine-comparison}. It can be seen that the PLCC and SROCC of the progressive method from coarse to fine are both higher than 0.9, while those of the random initialization method used are only about 0.5. The above results fully demonstrate the effectiveness of the progressive prediction mechanism.

\subsection{Overall Assessment Performance}
To test the performance of PKT-PCQA, we conducted experiments on the SJTU-PCQA, M-PCCD, and WPC datasets. We randomly selected about 80\% of the point clouds for training and the remaining 20\% for testing. There were no overlapping samples between the training and the testing sets. The performance results of the proposed network on different datasets are shown in Table~\ref{performance of different datasets}.  We should note that the PLCC and SROCC can be positive or negative, and the closer the absolute value of these to 1, the better the performance.
\begin{table}[t!]
\centering \caption{The performance of PKT-PCQA on different datasets}
\begin{tabular}{ccc}
\toprule \hline  Datasets    &PLCC     &SROCC\\\hline
\rowcolor{mygray} SJTU-PCQA~\cite{yang2020predicting}  &0.9122         &0.9316         \\
M-PCCD~\cite{alexiou2019comprehensive}  &0.8849         &0.9134         \\
\rowcolor{mygray} WPC~\cite{su2019perceptual}~\cite{liu2022perceptual}  &0.5603         &0.5571         \\\hline \bottomrule
\end{tabular}
\label{performance of different datasets}
\end{table}

The result in Table~\ref{performance of different datasets} shows that our network achieved excellent results on SJTU-PCQA and M-PCCD. Its performance was lower on WPC. One possible reason is that for WPC, the texture content of the point clouds is richer, and the content of the point clouds at each quality level is also richer, so the demand of data is also larger, and the amount of point cloud data in the WPC dataset can provide is very limited. For the proposed network, the task of predicting quality on WPC dataset is more difficult, therefore, the performance on WPC dataset degrades slightly.

\begin{table*}[t!]
\centering \caption{Accuracy comparison with existing FR, RR and NR point cloud quality metrics on different datasets}
\begin{tabular}{c|c|c|c|c|c|c|c}
\toprule
\hline
\multicolumn{2}{c|}{\diagbox[innerwidth=4cm]{Methods}{Datasets}}&\multicolumn{2}{c|}{SJTU-PCQA\cite{yang2020predicting}}&\multicolumn{2}{c|}{M-PCCD\cite{alexiou2019comprehensive}}&\multicolumn{2}{c}{WPC\cite{su2019perceptual}~\cite{liu2022perceptual}}\\ 
\hline Method Type  &Method &$\lvert PLCC \rvert$ &$\lvert SROCC \rvert$ &$\lvert PLCC \rvert$ &$\lvert SROCC \rvert$ &$\lvert PLCC \rvert$ &$\lvert SROCC \rvert$\\
\hline \multirow{8}*{FR}
&$PSNR_{MSE,p2po}$~\cite{mekuria2017performance} &0.87 &0.82 &0.84  &0.91  &0.37  &0.34 \\
&$PSNR_{HF,p2po}$~\cite{mekuria2017performance} &0.69 &0.70 &0.24  &0.53  &0.20  &0.20\\
&$PSNR_Y$~\cite{mekuria2017performance2} &0.83 &0.82 &\textbf{0.99}  &\textbf{0.99}  &0.48  &0.46\\
&$PCQM$~\cite{meynet2020pcqm} &0.85 &0.91 &0.76  &0.99  &0.68  &0.73\\
&$GraphSIM$~\cite{yang2020inferring} &0.72 &0.63 &0.98  &0.98  &0.47  &0.46\\
&$IW-SSIM_{p}$~\cite{wang2010information}  &\textbf{0.96} &\textbf{0.95} &0.85  &0.96  &\textbf{0.83}  &\textbf{0.82}\\
&$PointSSIM$~\cite{alexiou2020towards} &0.89 &0.88 &0.95  &0.94  &0.36  &0.23\\
&$LP-PCQM$~\cite{9448078} &0.93 &0.92 &0.99  &0.97  &0.50  &0.48 \\\cline{1-8}
RR &$PCM\_{RR}$~\cite{viola2020reduced} &0.60 &0.63 &0.87  &0.91  &0.36  &0.35 \\\cline{1-8}  
\multirow{2}*{NR}
&$PQA-NET$~\cite{liu2021pqa} &0.77 &0.71 &0.86  &0.85  &0.30  &0.20  \\
&\textbf{PKT-PCQA}&\color{blue}\textbf{0.91} &\color{blue}\textbf{0.93} &\color{blue}\textbf{0.88}  &\color{blue}\textbf{0.91}  &\color{blue}\textbf{0.56}  &\color{blue}\textbf{0.56} 
\\\hline
\bottomrule
\end{tabular}
\label{comparison in different metric}
\end{table*}
\subsection{Performance Comparison with other Methods}
We alse compared PKT-PCQA with standard and state-of-the-art FR and RR point cloud quality metrics on different datatsets. The compared metrics were chosen to cover a diversity of design philosophies, including point-based, feature-based, projection-based and deep learning-based. The FR quality metrics include $PSNR_{MSE,p2po}$, $PSNR_{HF,p2po}$, $PSNR_Y$, $PCQM$, $GraphSIM$, $IW-SSIM_{p}$, $PointSSIM$, and $LP-PCQM$. The RR quality metrics include $PCM\_{RR}$. More importantly, the sophisticated advanced $PQA-NET$~\cite{liu2021pqa} method is selected as the compared NR point cloud quality metric to prove the effectiveness of the proposed $PKT-PCQA$ method. To map the dynamic range of the scores from objective quality assessment models into a common scale, the logistic regression recommended by VQEG is used~\cite{logistic4}. 
These results are shown in Table.~\ref{comparison in different metric} and they provide some useful insights with respect to the approaches for point cloud quality assessment. First of all, the performance of RR and NR point cloud quality metrics are generally lower than that of the FR point cloud quality metrics. Obviously, the RR and NR metrics can only use part or no original information, while, the FR metric can take advantage of all the original information. Limited by the unfair usage of the original information, the performance of RR and NR methods are reasonably worse than the FR methods. From the results,we can see that the performance of the proposed PKT-PCQA is the best in the RR and NR quality metrics and only slightly lower than the best FR methods, indicating the advantage of the proposed PKT-PCQA. In the WPC dataset, which provides the largest number of samples in the three datasets, both PLCC and SROCC of the PKT-PCQA are 0.56 which is the best in the NR point cloud quality metrics. More surprisingly, the performance of PKT-PCQA was better than the RR quality metric $PCM_{RR}$, and is only slightly lower than the best FR quality metric $IW-SSIM_{p}$. The performance of PKT-PCQA was also impressive on SJTU-PCQA and M-PCCD datasets. Similarly, PKT-PCQA performed better than the other NR point cloud quality metric and even better than the RR point cloud quality metric. The PLCC and SROCC of PKT-PCQA were even larger than 0.91 on SJTU-PCQA. On the M-PCCD dataset, the SROCC reached 0.91. The SROCC and PLCC of the best FR point quality metric were 0.95 and 0.96, which are only 0.02 and 0.05 larger than that of PKT-PCQA. It is worth indicating that the NR point cloud quality metric PKT-PCQA does not need to refer to the original point cloud information, so its application scenarios are far more than the FR point cloud quality metrics.


\section{Conclusion}\label{sec:conclusion}
We proposed a novel no-reference quality assessment PKT-PCQA is proposed using coarse-to-fine progressive transfer based on human visual perception mechanism. First, the key clusters are extracted based on the global and local information. Then, to improve the similarity between prediction quality and subjective quality of the point cloud, an attention mechanism is incorporated into the network design. Next, a progressive quality assessment algorithm is designed to transfer the quality classification coarse-grained information to the quality prediction fined-grained information. Finally, multiple datasets are exploited to analyze the performance of the proposed algorithm. The experimental results show the effectiveness and superiority of the proposed method.

In the future work, we will develop an accuracy adaptive quality classification algorithm to accurately distinguish good, fair, and bad point cloud quality, so that more point cloud data without MOS can be used.

\section{Acknowledgement}
\label{sec:Acknowledgement}
This work was supported in part by Natural Sciences and Engineering Research Council of Canada, in part by the National Science Foundation of China under Grant (62222110, and 62172259), in part by the Taishan Scholar Project of Shandong Province (tsqn202103001), in part by Shandong Provincial Natural Science Foundation of China under Grants (ZR2022MF275, ZR2022QF076, ZR202206060017, ZR2018PF002, and ZR2021MF025), and in part by the Joint funding for smart computing of Shandong Natural Science Foundation of China under Grant ZR2019LZH002, and jointly completed by the State Key Laboratory of High Performance Server and Storage Technology, Inspur Group, Jinan.


\bibliographystyle{IEEEtran}
\bibliography{ref}
\end{document}